%% file: acl_latex.tex
\documentclass[11pt]{article}

\usepackage[preprint]{acl}

\usepackage{times}
\usepackage{latexsym}

\usepackage[T1]{fontenc}
\usepackage[utf8]{inputenc}

\usepackage{microtype}

\usepackage{inconsolata}

\usepackage{graphicx}
\usepackage{url}

\usepackage{booktabs,tabularx,array,xurl,hyperref}
\usepackage{multirow,placeins,adjustbox,colortbl}
\usepackage[most]{tcolorbox}
\usepackage{amsmath,amssymb}

\definecolor{yang}{HTML}{2da02c}

\definecolor{revcolor}{HTML}{C1440E}

\definecolor{revcolorb}{HTML}{0072B2}
\definecolor{AC}{RGB}{51,34,136}
\definecolor{AU}{RGB}{17,119,51}
\definecolor{TNG}{RGB}{68,170,153}
\definecolor{IE}{RGB}{136,204,238}
\definecolor{ST}{RGB}{221,204,119}
\definecolor{QU}{RGB}{204,102,119}
\definecolor{NE}{RGB}{170,68,153}
\definecolor{NS}{RGB}{136,34,85}
\definecolor{YW}{RGB}{255,204,204}
\definecolor{WP}{RGB}{229,204,255}

\newcommand{\colorcell}[1]{
    \pgfmathsetmacro{\value}{#1}

    \pgfmathparse{\value > \MaxNumber ? 1 : 0}
    \ifnum\pgfmathresult=1
        \hspace{-0.33em}\cellcolor{high!\opacity}{#1}
    \else
        \pgfmathparse{\value > \MidNumber ? 1 : 0}
        \ifnum\pgfmathresult=1
            \pgfmathparse{int(round(100*(\value - \MidNumber)/(\MaxNumber - \MidNumber)))}
            \xdef\tempa{\pgfmathresult}
            \hspace{-0.33em}\cellcolor{high!\tempa!yellow!\opacity}{#1}
        \else
            \pgfmathparse{int(round(100*(\MidNumber - \value)/(\MidNumber - \MinNumber)))}
            \xdef\tempa{\pgfmathresult}
            \hspace{-0.33em}\cellcolor{low!\tempa!yellow!\opacity}{#1}
        \fi
    \fi
}

\usepackage{twemojis}
\newcommand{\email}{\scalebox{1}{\twemoji{email}}}

\definecolor{purpleGreek}{HTML}{800080}
\definecolor{blueGreek}{HTML}{4891e7}
\definecolor{redGreek}{HTML}{d9381e}
\definecolor{greenGreek}{HTML}{088f8f}
\definecolor{yellowGreek}{HTML}{E0BC00}
\definecolor{orangeGreek}{HTML}{FFA500}

\newcommand{\uom}{{\color{purpleGreek}\boldsymbol{\pi}}}
\newcommand{\mc}{{\color{greenGreek}\boldsymbol{\kappa}}}
\newcommand{\sufe}{{\color{orangeGreek}\boldsymbol{\zeta}}}

\usepackage{microtype}
\usepackage[subtle]{savetrees}

\usepackage{silence}
\title{Calibrated Typed Decisions for Entity, Relation, and Event Extraction}

\author{
  \parbox{\textwidth}{\centering
    \textbf{Yang Wang}$\hspace{.1em}^{\uom}$,
    \textbf{Hanlin Xiao}$\hspace{.1em}^{\uom}$,
    \textbf{Yanan Ma}$\hspace{.1em}^{\mc}$,
    \textbf{Chenghao Xiao}$\hspace{.1em}^{\sufe}$,
    \textbf{Chenghua Lin}$\hspace{.1em}^{\uom}$$\hspace{.1em}^{\email}$
  \\ \vspace{0.4cm}
    \textnormal{$\hspace{.1em}^{\uom}$University of Manchester},
    \textnormal{$\hspace{.1em}^{\mc}$Mayo Clinic},
    \textnormal{$\hspace{.1em}^{\sufe}$Shanghai University of Finance and Economics}
  }
}

\begin{document}
\maketitle

\newcommand{\corresponding}[1]{
  \begingroup
  \renewcommand{\thefootnote}{}
  \makeatletter
  \def\@makefnmark{}
  \makeatother
  \begin{NoHyper}
    \footnotetext{#1}%
  \end{NoHyper}
  \endgroup
}
\corresponding{\email\hspace{.1em} The corresponding author is Chenghua Lin, who can be contacted at \url{chenghua.lin@manchester.ac.uk}.}

\begin{abstract}
% Generalising information extraction across datasets requires models to handle changes in domain and annotation guidelines, even when the requested types are familiar. We present \textsc{CalOn}, a unified 0.6B-parameter encoder that formulates entity, relation, and event extraction as probabilistic decisions over input spans and span pairs. Its central design is a shared decision rule. Queried types and candidates interact, and each candidate makes a single choice among the types and a null class. It thresholds the total probability of the queried types to decide whether to extract a candidate, then assigns the most probable type. Training uses diverse sources, including synthetic data, distinguishes unannotated types from negatives, and combines supervised learning with rewards on sampled decisions. A closed-form correction accounts for the probability bias caused by down-weighting null candidates. We evaluate on 40 benchmarks covering conventional information extraction and tasks represented as span-pair extraction, including aspect sentiment triplet extraction, pronoun resolution, coreference, and temporal and causal relations between events. \textsc{CalOn} achieves higher F1 than the evaluated baselines on most benchmarks. Ablations show that jointly choosing among types and null improves extraction across all three tasks. The correction and temperature refitting reduce average calibration error for entities, relations, and event arguments with little change in F1. 
% We release our code, model, and dataset at \url{https://anonymous.4open.science/r/970D}.
Information extraction models often perform well on the datasets they are trained on, but transferring to new domains and annotation guidelines remains challenging. Supporting user-specified types alone does not resolve this problem. We present \textsc{CalOn}, a unified 0.6B-parameter encoder for entity, relation, and event extraction across datasets and schemas. The model jointly encodes text and type definitions, then directly identifies entity spans, relations between spans, and event triggers and arguments. Its extraction decisions consider the requested types together, using their combined evidence to identify relevant candidates before assigning a type. This gives all three tasks a common probabilistic formulation while keeping every extracted mention anchored to the input text. We train on diverse sources, and supplement supervised learning with a reward that encourages correct extractions and penalises incorrect ones. Across 40 benchmarks, \textsc{CalOn} outperforms the evaluated baselines on most benchmarks. It leads all reported relation extraction benchmarks and achieves the highest average F1 for event triggers and arguments with gold triggers. The same model also performs strongly on aspect sentiment triplet extraction and pronoun resolution through its span-pair formulation. A correction for the probability bias introduced during training improves confidence estimates for entities, relations, and event arguments after calibration, with little change in F1.
\end{abstract}

\section{Introduction}
\label{sec:introduction}

Information extraction (IE) converts plain text into structured records of entities, relations, and events. The required information varies across applications, so an extractor must follow a schema specifying the types to identify. 
For example, job-description analysis requires identifying skills, qualifications, and experience \citep{green-etal-2022-development}. 
Traditional systems are trained for a fixed set of types, while recent models accept type names, descriptions, or annotation guidelines as part of the input \cite{sainz2024gollie, zaratiana2025gliner2}. This allows users to specify what to extract without training a separate model for every schema. The field has therefore moved towards general-purpose extraction, with a single model intended to serve multiple tasks and datasets \cite{lu-etal-2022-unified, wang2023instructuie}.

% Information extraction (IE) identifies entities, relations, and events in text according to a specified schema. A general-purpose extractor should support new datasets without labelled examples from each one, and even familiar labels become difficult when the domain, writing style, or annotation guidelines change \citep{bassignana-plank-2022-crossre,sainz2024gollie}. 
% We study zero-shot transfer across datasets, allowing types to overlap with those seen during training. 
% The challenge is how to make a compact extractor transfer across these changes while keeping precise spans and explicit control over which predictions it returns.

Supporting a new schema does not by itself ensure accurate extraction on a new dataset. Models must transfer beyond the domains and annotation conventions represented in their training data \citep{bassignana-plank-2022-crossre,sainz2024gollie}. Discriminative models predict spans and their types directly from the input. UniEX \citep{ping-etal-2023-uniex}, GLiNER \citep{zaratiana-etal-2024-gliner}, GLiNER2 \citep{zaratiana2025gliner2}, and GLiFormer \citep{Knowledgator2026GLiFormer} support flexible schemas, but their performance varies across domains and extraction tasks, as our evaluation shows. Generative models such as UIE \citep{lu2022unified}, InstructUIE \citep{wang2023instructuie}, and GoLLIE \citep{sainz2024gollie} express extraction targets through text generation. This accommodates detailed instructions, but generated outputs can introduce unsupported entities or fail to preserve the original mention boundaries. GPT-NER reports false entity predictions \citep{wang-etal-2025-gpt}, while GNER documents boundary errors and omissions, additions, or changes to input words \citep{ding2024rethinking}. 
% Prompting strategies that work well for one IE task may not transfer to another \citep{huang2026reevaluating}. 
The remaining challenge is to support user-specified schemas while maintaining extraction accuracy across domains and annotation guidelines.

% Discriminative extractors predict directly over the input. UniEX \citep{ping-etal-2023-uniex} unifies extraction through span and token-pair predictions, and GLiNER \citep{zaratiana-etal-2024-gliner}, GLiNER2 \citep{zaratiana2025gliner2}, and GLiFormer \citep{Knowledgator2026GLiFormer} offer flexible schema-based interfaces, which do not by themselves ensure accuracy across specialised domains or across the three IE tasks. Generative approaches such as UIE \citep{lu2022unified}, InstructUIE \citep{wang2023instructuie}, and GoLLIE \citep{sainz2024gollie} formulate information extraction as text generation, allowing the target schema to be specified through prompts. However, generated predictions may contain false entities or fail to preserve the exact wording and boundaries of mentions in the input. The GPT-NER study \citep{wang-etal-2025-gpt} identifies a tendency to label non-entity text as entities. The GNER study \citep{ding2024rethinking} documents boundary errors and cases where generation omits, adds, or changes input words. In few-shot experiments, LLMs prompted without IE-specific fine-tuning also underperform fine-tuned smaller models in most evaluated settings \citep{ma-etal-2023-large}. Training coverage also matters, as zero-shot NER depends on the similarity and frequency of training types \citep{golde2025familiarity}.

To address this challenge, we investigate whether reward-based training can improve a schema-conditioned extractor beyond likelihood supervision alone. We present \textsc{CalOn}, a unified 0.6B-parameter encoder for entity, relation, and event extraction. Drawing on RLCD's emphasis on decisions with explicit probabilities \citep{typesafe2026rlcd}, we supplement supervised learning with optimisation of sampled extraction decisions. As in GRPO \cite{shao2024deepseekmath}, multiple predictions for the same input provide a reward baseline without requiring an additional critic. For each candidate span or span pair, we sample predictions by adding Gaussian noise to the type scores. Correctly typed extractions receive positive rewards, false positives and incorrect types receive negative rewards, and predictions of no extraction receive zero reward. We compare each prediction's reward with the average reward of the other samples and use this difference to reinforce higher-reward decisions. The resulting objective combines likelihood supervision with direct optimisation of the expected reward of sampled extractions. 
The motivation is to strengthen correct predictions when the model can also produce incorrect alternatives. For the same candidate, changing the type scores can lead to a correct extraction, a wrong type, or a false positive. Comparing their rewards gives an additional training signal that favours the correct extraction over these mistakes. We use this signal to supplement likelihood supervision, aiming to make the model less likely to return an incorrect prediction when its scores change.

Across 40 benchmarks, \textsc{CalOn} outperforms the evaluated baselines on most benchmarks. It leads every reported relation extraction benchmark, including tasks represented as span-pair extraction such as aspect sentiment triplet extraction, pronoun resolution, coreference, and temporal and causal relations between events. It also achieves the highest average F1 for event triggers and arguments with gold triggers. These results demonstrate that a single extraction model can transfer across datasets and support tasks beyond conventional entity, relation, and event extraction. We make the following contributions.
\begin{enumerate}
    \item We introduce \textsc{CalOn}, a unified encoder for extraction under user-specified schemas. It handles entities, relations, and events through predictions over spans and span pairs. The same formulation also supports tasks beyond conventional IE (\S\ref{sec:methodology}).
    \item We develop a training objective that combines likelihood supervision with relative rewards on sampled extraction decisions. Gaussian perturbations generate predictions for each candidate, and comparison of their rewards guides the model towards correct extractions over false positives and incorrect type assignments, without an additional critic (\S\ref{sec:loss}).
\end{enumerate}

\section{Related Work}
\label{sec:related_work}

\paragraph{Extraction across schemas and data sources.}
UIE \citep{lu2022unified} unifies entity, relation, and event extraction through structured generation, while USM \citep{lou2023universal} and UniEX \citep{ping-etal-2023-uniex} use schema-guided matching. InstructUIE \citep{wang2023instructuie} uses instruction tuning, and GoLLIE \citep{sainz2024gollie} incorporates annotation guidelines to support generative extraction under user-specified schemas. GLiNER \citep{zaratiana-etal-2024-gliner} and GLiREL \citep{boylan-etal-2025-glirel} match spans and span pairs, respectively, to requested types. GLiNER2 \citep{zaratiana2025gliner2} and GLiFormer \citep{Knowledgator2026GLiFormer} support multiple tasks within one model. 
Training resources have also expanded through LLM annotations and synthetic examples, as in UniversalNER \citep{zhou2024universalner}, FiNERweb \citep{golde-etal-2026-finerweb}, and GUIDEX \citep{fuente-etal-2025-guidex}. 
% Synthetic supervision has also been used to train generative models for medical text anonymisation \citep{xiao2026generalizable}. 
Combining sources requires handling differences in annotation coverage, since an unlabelled span is not necessarily a negative example \citep{jie-etal-2019-better, li2021empirical}. 
% When a type's presence is unknown in a training example, \textsc{CalOn} excludes that type from the training loss.

\paragraph{Learning extraction decisions.}
Span-based IE models use a null class to reject non-entity spans \citep{eberts2020span, zhong-chen-2021-frustratingly}. For generative IE, LA-RL \citep{you2026rl} applies reinforcement learning through label-aware self-reflection. Inspired by RLCD \citep{typesafe2026rlcd}, \textsc{CalOn} adds a reward term for sampled span and span-pair predictions. Correctly typed extractions receive positive rewards, while false positives and incorrect type assignments receive negative rewards.

\paragraph{Calibration.}
Temperature scaling can improve the calibration of neural-network predictions after training \citep{guo2017caibration}. 
% However, post-hoc calibration can remain unreliable under distribution shift \citep{ovadia2019can}. 
Calibration can also be improved through training, with label smoothing helping pre-trained Transformers on out-of-domain data \citep{desai-durrett-2020-calibration}. Even when probabilities are calibrated, selecting a threshold to maximise F1 remains dependent on the distribution of prediction scores, as shown for binary classification by \citet{lipton2014optimal}. Class-dependent sampling can introduce a separate bias in probability estimates. Corrections for this effect have been studied in rare-event logistic regression \citep{king2001logistic} and click prediction \citep{he2014practical}. \textsc{CalOn} applies this correction principle to the lower loss weight assigned to null candidates in span and span-pair extraction.

\begin{figure*}
    \centering
    \includegraphics[width=1.0\linewidth]{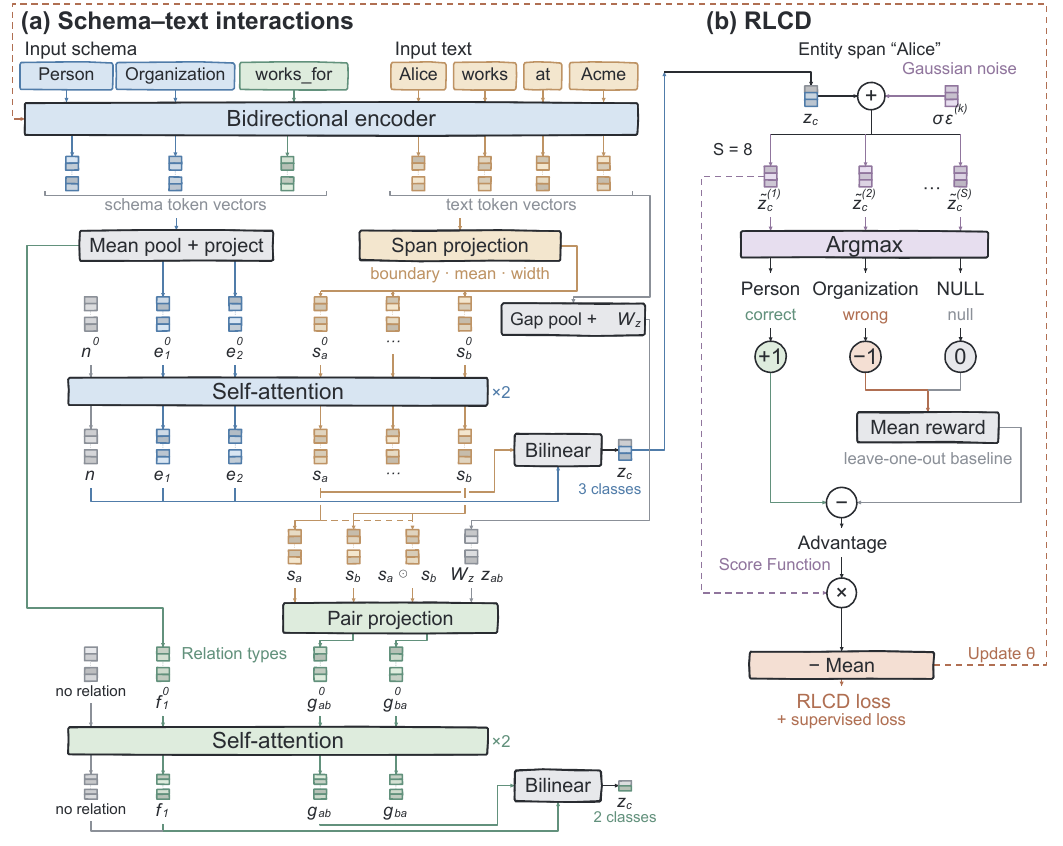}
    \caption{Overview of \textsc{CalOn}: (a) schema-text encoding with type-candidate interaction and bilinear scoring over the queried types and a null class, and (b) the RLCD reward on sampled typed decisions added to the supervised loss.}
    \label{fig:placeholder}
\end{figure*}

\section{Methodology}
\label{sec:methodology}

\textsc{CalOn} jointly encodes the input text and a schema of entity and relation types with optional descriptions (\S\ref{sec:encoder}). For each candidate span or ordered pair of entity spans, it predicts a probability distribution over the queried types and a null class (\S\ref{sec:decisions}). The training objective combines a negative log-likelihood loss on the gold labels with a reward term based on sampled decisions (\S\ref{sec:loss}). At inference, the model selects candidates by thresholding the total probability assigned to the queried types, then labels each selected candidate with the most probable queried type (\S\ref{sec:decoding}). Figure~\ref{fig:placeholder} illustrates this formulation.

\subsection{Encoder}
\label{sec:encoder}

For entity types $\{(y_m, d_m)\}_{m=1}^{M}$ with descriptions $d_m$, relation types $\{(r_k, d'_k)\}_{k=1}^{K}$ and a text $x$, the input is
\begin{equation}
\begin{aligned}
&\texttt{<LABEL>}\, y_1{:}\, d_1 \;\cdots\; \texttt{<LABEL>}\, y_M{:}\, d_M \\
&\texttt{<REL>}\, r_1{:}\, d'_1 \;\cdots\; \texttt{<REL>}\, r_K{:}\, d'_K \;\; \texttt{<TEXT>}\, x
\end{aligned}
\label{eq:input}
\end{equation}
where \texttt{<LABEL>}, \texttt{<REL>} and \texttt{<TEXT>} are added special tokens and a type without a description is given by its name.
The encoder, Qwen3 \citep{qwen3} with its causal mask removed, gives token representations $\mathbf{H} \in \mathbb{R}^{T \times D}$; the text is tokenised separately so that each token keeps its character offsets.
A candidate span $(i, j)$ of at most $W$ text tokens is represented by its first and last token, the mean $\bar{\mathbf{h}}_{ij}$ of its tokens, and a width embedding,
\begin{equation}
\mathbf{s}^{0}_{ij} = \mathbf{W}_s\big[\mathbf{H}_i ; \mathbf{H}_j ; \bar{\mathbf{h}}_{ij} ; \mathbf{w}_{j-i}\big]
\label{eq:span}
\end{equation}
An entity type is represented by $\mathbf{e}^{0}_m = \mathbf{W}_e\bar{\mathbf{h}}_{m}$, where $\bar{\mathbf{h}}_{m}$ is the mean over the tokens of its name and description, and a relation type by $\mathbf{f}^{0}_k = \mathbf{W}_r\bar{\mathbf{h}}'_{k}$; all projections output vectors of size $A$.

\paragraph{Interaction layers.}
The score of a type for a span should depend on the other queried types and on overlapping candidates, so the types and candidates of an instance attend to each other. The sequence $[\mathbf{n}^{0}; \mathbf{e}^{0}_1; \ldots; \mathbf{e}^{0}_M; \mathbf{s}^{0}_1; \ldots; \mathbf{s}^{0}_C]$, where $\mathbf{n}^{0}$ is a learned vector for the null class and each element is added to a learned embedding of its kind (null, type or span), passes through $L$ pre-norm Transformer blocks with full self-attention and no position information, which give $\mathbf{n}$, $\mathbf{e}_m$ and $\mathbf{s}_c$.

\subsection{Typed Decisions with a Null Class}
\label{sec:decisions}

\paragraph{Label status.}
Training sources differ in what a missing annotation means. In an exhaustively annotated dataset such as Few-NERD \cite{ding-etal-2021-nerd}, a queried type with no mention in a sentence is absent; in a partially annotated one, such as RAMS \cite{ebner-etal-2020-multi}, which labels one event per passage, it may simply not have been annotated. Each queried type in a training instance therefore has one of four statuses: \emph{positive} (it has a gold mention), \emph{negative} (the source states that it does not occur, e.g.\ types listed as absent when a synthetic text is generated), \emph{closed-world negative} (absent because the source annotates its types exhaustively), or \emph{unknown} (its absence cannot be inferred). Both kinds of negative are trained as absent, whereas unknown types are left out of the decision; $\mathcal{K}$ denotes the queried types whose status is not unknown.

\paragraph{Entities.}
Each candidate span $c$ takes exactly one class from $\{0\} \cup \mathcal{K}$, where $0$ is the null class, with logits
% \begin{align}
% z_{m,c} &= (\mathbf{V}\mathbf{e}_m)^{\top}(\mathbf{U}\mathbf{s}_c) / \sqrt{r} + b, \label{eq:logit}\\
% z_{0,c} &= (\mathbf{V}\mathbf{n})^{\top}(\mathbf{U}\mathbf{s}_c) / \sqrt{r}, \label{eq:null}
% \end{align}
\begin{align}
z_{m,c} &= \frac{1}{\sqrt{r}} (\mathbf{V}\mathbf{e}_m)^{\top}(\mathbf{U}\mathbf{s}_c) + b \label{eq:logit}\\
z_{0,c} &= \frac{1}{\sqrt{r}} (\mathbf{V}\mathbf{n})^{\top}(\mathbf{U}\mathbf{s}_c) \label{eq:null}
\end{align}
where $\mathbf{U}, \mathbf{V} \in \mathbb{R}^{r \times A}$ form a bilinear form of rank $r$, $b$ is a learned bias, and the null logit depends on the span through $\mathbf{n}$. The class distribution is
\begin{equation}
% p_c(m) = \frac{\exp z_{m,c}}{\sum_{m' \in \{0\} \cup \mathcal{K}} \exp z_{m',c}} .
p_c(m) = \frac{\exp z_{m,c}}{\sum\limits_{m' \in \{0\} \cup \mathcal{K}} \exp z_{m',c}}
\label{eq:choice}
\end{equation}
If an instance has an unknown type, spans without a gold type are ignored in training, since their class may be that type.

\paragraph{Relations.}
Relation candidates are the ordered pairs $(a, b)$, $a \neq b$, of entity spans in an instance (at most 24 spans; in training the gold entity spans, those in a gold relation first). A pair is represented by the span vectors, their element-wise product, and the mean $\mathbf{z}_{ab}$ of the tokens between the two spans (zero if there are none),
\begin{equation}
\mathbf{g}^{0}_{ab} = \mathbf{W}_g\big[\mathbf{s}_a ; \mathbf{s}_b ; \mathbf{s}_a \!\odot\! \mathbf{s}_b ; \mathbf{W}_z \mathbf{z}_{ab}\big]
\label{eq:pair}
\end{equation}
Each pair makes the same single choice among ``no relation'' and the queried relation types (Eqs.~\eqref{eq:logit}-\eqref{eq:choice}), after interaction layers of its own over the no-relation vector, the relation types $\mathbf{f}^{0}_k$ and the pairs, with a separate bilinear form, bias and null vector. Since $(a, b)$ and $(b, a)$ are different candidates, the model also predicts the direction of a relation.

\paragraph{Events.}
An event mention is a trigger span of an event type with argument spans that fill its roles, and each event text gives two training instances. In the first, the event types are queried as entity types and the triggers are their spans. In the second, the event types present in the text are queried as entity types, with their triggers as spans, and their roles both as entity types, labelling the argument spans, and as relation types; each argument is a relation from its trigger to its span, labelled with its role. At inference, the entity decision first predicts triggers; a second pass over the predicted event types and their roles proposes argument spans, and the relation decision labels each (trigger, span) pair with a role of the trigger's event type.

\subsection{Training Objective}
\label{sec:loss}

Let $\mathcal{G}_c \subseteq \mathcal{K}$ be the gold types of candidate $c$ (a span can have a coarse and a fine type), with $\mathcal{G}_c = \{0\}$ if it has none. We maximise their total probability, $\ell_c = -\log \sum_{g \in \mathcal{G}_c} p_c(g)$ (Appendix~\ref{sec:appendix_method} gives an optional Brier term). Nearly all candidates belong to the null class, so they get a weight $\omega \le 1$, and the loss of an instance is the weighted mean
\begin{equation}
% \mathcal{L}_{\mathrm{inst}} = \textstyle\sum_{c} \omega_c\, \ell_c \big/ \sum_{c} \omega_c ,
\mathcal{L}_{\mathrm{inst}} = \frac{\textstyle\sum\limits_{c} \omega_c\, \ell_c}{\sum\limits_{c} \omega_c}
\label{eq:inst}
\end{equation}
averaged over the instances of a batch to give $\mathcal{L}_{\mathrm{ent}}$.

\paragraph{Decision reward.}
Each candidate also acts as a stochastic policy. Its logits $\mathbf{z}_c$ over $\{0\} \cup \mathcal{K}$ are perturbed, $\tilde{\mathbf{z}}^{(k)}_c = \mathbf{z}_c + \sigma \boldsymbol{\epsilon}^{(k)}$ for $k = 1, \ldots, S$, and each sample takes the decision $\arg\max_m \tilde{z}^{(k)}_{m,c}$, rewarded as the evaluation scores it: $R = 1$ for a gold type, $R = -1$ for a type that is not gold, and $R = 0$ for the null class. With the advantage $\delta^{(k)}_c = R^{(k)}_c - \bar{R}^{(-k)}_c$ over the mean reward of the other samples,
\begin{equation}
\ell^{\mathrm{PG}}_c = -\frac{1}{S}\sum_{k=1}^{S} \delta^{(k)}_c \log \pi\big(\tilde{\mathbf{z}}^{(k)}_c \mid \mathbf{z}_c\big)
\label{eq:pg}
\end{equation}
with $\pi(\cdot \mid \mathbf{z}_c) = \mathcal{N}(\mathbf{z}_c, \sigma^2 \mathbf{I})$, is the score-function estimate of the negative expected reward; it is averaged with the weights of Eq.~\eqref{eq:inst} and added with weight $\rho$. The relation loss $\mathcal{L}_{\mathrm{rel}}$ is computed in the same way over candidate pairs, with its own null weight, and $\mathcal{L} = \mathcal{L}_{\mathrm{ent}} + \lambda\, \mathcal{L}_{\mathrm{rel}}$.

\subsection{Decoding and Calibration}
\label{sec:decoding}

A span is extracted if $1 - p_c(0) \ge \tau$ and labelled with $\arg\max_{m \in \mathcal{K}} p_c(m)$; relations use the same rule, and overlapping spans are allowed. $1 - p_c(0)$ is the probability that the span belongs to any queried type, whereas the predicted type's probability also depends on how the schema divides that mass among overlapping types (e.g., a coarse and a fine type), so thresholding it would reject spans whose evidence is shared. We therefore threshold the null class. For entities, the logits are divided by a temperature $T$ that minimises the multi-class Brier score, and $\tau$ maximises F1, both on the calibration splits of the training sources. For relations, $T = 1$ and $\tau$ is fitted on REBEL \cite{huguet-cabot-navigli-2021-rebel-relation} with gold entities.

\paragraph{Prior correction.}
Down-weighting null candidates by $\omega$ (Eq.~\eqref{eq:inst}) shifts the minimiser of the loss. The model reports the odds of the null class against the types $\omega$ times too low, as after down-sampling negatives \cite{king2001logistic, he2014practical}. The correction adds $\log(1/\omega)$ to the null logit before the division by $T$, leaving the relative odds of the types unchanged. It needs no data and does not change the ranking of candidates at a fixed $T$, so after refitting $T$ and $\tau$ it changes mainly the calibration (\S\ref{sec:results_and_discussion}).

\input{tables/ner_with_description}
\input{tables/re_with_description}
\input{tables/ee_with_description}

\section{Experimental Setting}
\label{sec:experimental_setting}

\paragraph{Training data.}
\label{sec:training_data}
We train on 3.1M instances from 84 sources covering entities, relations, and events: human-labelled data, LLM-annotated text, and synthetic examples spanning diverse domains, genres, and schemas.
% We check synthetic annotations with additional models, remove near-duplicate texts, and filter training texts that overlap substantially with evaluation sentences. \textsc{CalOn} uses no labelled examples from any target dataset for training, model selection, or calibration. Appendix~\ref{sec:appendix_data_preparation} details data preparation and sampling; Appendix~\ref{sec:appendix_data} describes the type distribution and domain coverage.
We check synthetic annotations with additional models, remove near-duplicate texts, and filter training texts that substantially overlap with evaluation sentences. The training-data generation pipeline does not use evaluation-set texts or annotations as inputs, demonstrations, or style references. No labelled examples from any target dataset are used for training, model selection, or calibration.
Appendices~\ref{sec:appendix_data_preparation} and~\ref{sec:appendix_data} describe the preparation, sampling, type distribution and domain coverage.

\paragraph{Evaluation.}
\label{sec:evaluation}
% We evaluate on 40 benchmarks covering NER, RE, and EE across multiple domains. Appendix~\ref{sec:appendix_evaluation_details} describes the datasets and scoring. We report exact-match micro F1 for each benchmark. Each benchmark receives equal weight when computing average F1. The main results use type descriptions for models that support them. Appendix~\ref{sec:appendix_wo_desc} reports results with type names alone. Relation predictions must also match the gold direction. For RE, we report end-to-end F1 and F1 with gold entity spans. For EE, we report trigger classification F1 and argument classification F1 with gold and predicted triggers.
We evaluate on 40 benchmarks covering NER, RE, and EE across multiple domains. Appendix~\ref{sec:appendix_evaluation_details} describes the datasets and scoring. We report exact-match micro F1 for each benchmark. Each benchmark receives equal weight when computing average F1. The main results use type descriptions for models that support them. Appendix~\ref{sec:appendix_wo_desc} reports results with type names alone. Directed relation predictions must also match the gold direction. For RE, we report end-to-end F1 and F1 with gold entity spans. For EE, we report trigger classification F1 and argument classification F1 with gold and predicted triggers.

\paragraph{Baselines.}
\label{sec:baselines}
We compare with GLiNER2-large \citep{zaratiana2025gliner2}, GLiFormer-large \citep{Knowledgator2026GLiFormer}, GNER-LLaMA-7B \citep{ding2024rethinking} (NER only), InstructUIE \citep{wang2023instructuie}, and GoLLIE-7B \citep{sainz2024gollie}, and for RE with gold entities GLiREL-large \citep{boylan-etal-2025-glirel}, with default decoding and the same scoring code. GLiFormer, GNER, InstructUIE and GLiREL receive type names only, following their released interfaces; adding descriptions led to invalid outputs for GNER and InstructUIE. For RE with gold entities, we match the other baselines' end-to-end relation endpoints to gold entities by span or mention text and discard unmatched relations. GLiREL receives gold entities directly (Appendix~\ref{sec:appendix_baseline_details}). 
% We compare with GLiNER2-large \citep{zaratiana2025gliner2}, GLiFormer-large \citep{Knowledgator2026GLiFormer}, GNER-LLaMA-7B \citep{ding2024rethinking} (NER only), InstructUIE \citep{wang2023instructuie}, and GoLLIE-7B \citep{sainz2024gollie}, and for RE with gold entities GLiREL-large \citep{boylan-etal-2025-glirel}, with default decoding and the same scoring code. GLiFormer, GNER, InstructUIE and GLiREL receive type names only, following their released interfaces; adding descriptions led to invalid outputs for GNER and InstructUIE. For RE with gold entities, we keep the end-to-end relations of the other baselines whose two spans are gold entities, a stricter proxy that still depends on finding the spans; GLiREL is given the gold entities directly (Appendix~\ref{sec:appendix_baseline_details}). 
% InstructUIE's released training configuration includes MIT Movie, MIT Restaurant, and CASIE, and GoLLIE was trained on TACRED, whose sentences Re-TACRED reuses; these scores are marked $^\ddagger$.

\paragraph{Implementation.}
\label{sec:implementation}
We use Qwen3-0.6B \citep{qwen3} as our backbone to limit the computational cost of training and inference. We remove its causal mask so that text and schema tokens can attend to the full input \citep{li2023label}. We select the checkpoint, hyper-parameters, temperature, and extraction thresholds using only validation data from the training sources. Appendix~\ref{sec:appendix_implementation_details} provides further details.

\section{Results and Discussion}
\label{sec:results_and_discussion}

Tables~\ref{tab:ner}, \ref{tab:re}, and \ref{tab:ee} report results for entity, relation, and event extraction. We first discuss performance across tasks and domains, then examine the model components, calibration, and efficiency. Appendix~\ref{sec:appendix_results} provides further discussion of individual datasets.

\subsection{Main Results}

\paragraph{Entity extraction.}
\textsc{CalOn} achieves the highest average F1 on CrossNER, specialised entity extraction, and personal information extraction (Table~\ref{tab:ner}). Its largest gains are on scientific, biomedical, and measurement text. GLiFormer leads on CrossNER Music and Science, while InstructUIE leads on MIT Restaurant. InstructUIE's Restaurant score raises its MIT average above \textsc{CalOn}'s, and both MIT datasets were included in training for the evaluated InstructUIE checkpoint (Appendix~\ref{sec:appendix_baseline_details}).
% \textsc{CalOn} achieves the highest average F1 on CrossNER, specialised entity extraction, and personal information extraction (Table~\ref{tab:ner}). Its largest gains are on scientific, biomedical, and measurement text. GLiFormer leads on CrossNER Music and Science, while InstructUIE leads on MIT Restaurant. InstructUIE's Restaurant score raises its MIT average above \textsc{CalOn}'s, and both MIT datasets were included in training for the evaluated InstructUIE checkpoint (\S\ref{sec:baselines}).

\paragraph{Relation extraction.}
\textsc{CalOn} leads every reported RE benchmark, both with gold entities and where end-to-end scoring is available (Table~\ref{tab:re}). The same span-pair formulation also performs well on ASTE and pronoun resolution. End-to-end results are omitted for datasets that annotate relations only for selected entity pairs. GLiREL is evaluated only with gold entities because it does not support end-to-end extraction. 
We also compare GLiREL under two candidate-pair settings (Appendix~\ref{sec:appendix_baseline_details}). When only pairs with an annotated relation are supplied, its average relation typing accuracy across the 18 RE benchmarks is 40\%, compared with 79\% for \textsc{CalOn}. When every pair of gold entities is considered, their average F1 scores are 2.6 and 49.8, respectively. The latter setting tests both relation typing and rejection of unrelated pairs, highlighting an important requirement for relation extraction.

\paragraph{Event extraction.}
\textsc{CalOn} achieves the highest average F1 for trigger classification and argument classification with gold triggers (Table~\ref{tab:ee}). InstructUIE's CASIE score raises its average end-to-end argument F1 above \textsc{CalOn}'s. The CASIE training split was included in training for the evaluated InstructUIE checkpoint (Appendix~\ref{sec:appendix_baseline_details}). Excluding CASIE, \textsc{CalOn} achieves the highest average F1 for all three EE metrics, with end-to-end argument F1 of 10.9 compared with InstructUIE's 8.4. Individual dataset exceptions, including M2E2 and SPEED++ zika, are discussed in Appendix~\ref{sec:appendix_results}.
% \textsc{CalOn} achieves the highest average F1 for trigger classification and argument classification with gold triggers (Table~\ref{tab:ee}). InstructUIE's CASIE score raises its average end-to-end argument F1 above \textsc{CalOn}'s. The CASIE training split was included in training for the evaluated InstructUIE checkpoint (\S\ref{sec:baselines}). Excluding CASIE, \textsc{CalOn} achieves the highest average F1 for all three EE metrics, with end-to-end argument F1 of 10.9 compared with InstructUIE's 8.4. Individual dataset exceptions, including M2E2 and SPEED++ zika, are discussed in Appendix~\ref{sec:appendix_results}.

\subsection{Schemas and Type Descriptions}

Type descriptions are particularly helpful when relation names leave their meaning unclear. On CrossRE, descriptions increase average F1 with gold entities by about 16 points, while NER group averages change by one to three points. Descriptions can therefore provide information that the type names alone do not convey. 
Every entity type evaluated also occurs by name in our training data. The NER benchmarks thus test transfer across domains and annotation guidelines (Table~\ref{tab:familiarity}, Appendix~\ref{sec:appendix_familiarity}). We find no consistent increase in per-type F1 with how often a type is annotated during training. \textsc{CalOn}'s largest margin over the strongest baseline is observed for the least frequently annotated third of the entity types. These findings suggest that frequent type names alone do not explain its performance.

\subsection{Ablation Study}

\begin{figure*}[t]
\centering
\includegraphics[width=\textwidth]{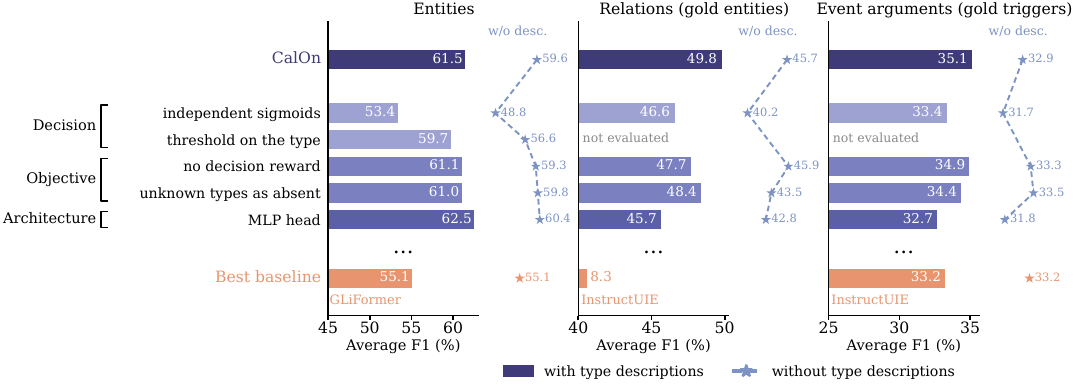}
\caption{Ablation results for \textsc{CalOn}. Each row changes one model component. Bars show average micro F1 (\%) with type descriptions over 15 NER benchmarks, 18 RE benchmarks with gold entities, and 7 EE benchmarks with gold triggers. Stars ($\star$) show results with type names alone on their own scales. The bottom row shows the strongest baseline for each task, averaged over the same benchmarks. InstructUIE's training data include some of the event benchmarks.}
\label{fig:ablation}
\end{figure*}

\paragraph{Extraction and type prediction.}
Independent sigmoid scores reduce average NER F1 from 61.5 to 53.4, the largest NER decrease among the tested ablations (Figure~\ref{fig:ablation}). Thresholding the most probable type also lowers NER F1. In the softmax model, related types can share a candidate's probability. A threshold on an individual type may therefore reject a candidate even when its total non-null probability is high. Thresholding the total non-null probability uses the evidence across types to decide whether to extract. These results support learning a joint distribution over types and a null class while separating extraction from type selection.

\paragraph{Architecture and training objectives.}
The MLP head variant slightly improves NER but lowers RE and event-argument F1. This suggests that the full prediction head is more useful for span-pair prediction than for entity typing. With type descriptions, the decision reward mainly benefits RE, and treating unknown types as negatives lowers F1 across all three tasks. These training effects are smaller and less consistent for NER and EE without descriptions, suggesting that their benefits can depend on how the schema is presented.

\subsection{Calibration}

Null-class correction addresses the probability bias introduced by giving null candidates less weight in the training loss. After correction and refitting, average expected calibration error falls from 0.25 to 0.10 for relations, from 0.33 to 0.23 for entities, and from 0.51 to 0.39 for event arguments, with little change in F1 (Table~\ref{tab:calibration}). The results indicate more accurate confidence estimates with similar extraction accuracy. Event predictions reuse the entity and relation calibration parameters. Their calibration also improves, although event arguments retain the highest average calibration error.

\subsection{Efficiency}

\input{tables/efficiency}

For a given schema, \textsc{CalOn} scores all queried types and candidates in one forward pass. Under the same PyTorch setting, it processes 120 sentences per second, 5 to 25 times the throughput of the baselines (Table~\ref{tab:efficiency}, Appendix~\ref{sec:appendix_efficiency}). This advantage includes GLiNER2 and GLiFormer, which have similar parameter counts. \textsc{CalOn} also achieves the highest average NER F1 among the compared models. Its peak GPU memory usage is 8.0 GB, higher than GLiNER2 and GLiFormer, so the throughput advantage is accompanied by greater runtime memory use.

\section{Conclusion}
\label{sec:conclusion}

We presented \textsc{CalOn}, a unified 0.6B-parameter encoder for extraction under user-specified schemas. The model handles entities, relations, and events through span and span-pair predictions. Its training objective combines likelihood supervision with relative rewards on sampled extractions, favouring correct predictions over false positives and incorrect type assignments. Across 40 benchmarks, \textsc{CalOn} outperforms the evaluated baselines on most benchmarks and leads every reported relation extraction benchmark. It also achieves the highest average F1 for event triggers and arguments with gold triggers. Its advantages extend to tasks represented as span-pair extraction, including aspect sentiment triplet extraction, pronoun resolution, coreference, and temporal and causal relations between events. Probability correction and calibration improve confidence estimates across all three extraction tasks with little change in F1. 
% These results demonstrate the breadth of a single extraction model that combines direct span prediction, reward-based training, and explicit probabilities.

\section*{Limitations}

Our training data and evaluation are limited to English, and transfer to other languages has not been tested. Candidate spans are capped at 32 tokens, and relations are predicted only between spans within the same input. These limits leave longer mentions and relations across inputs outside the model's current scope. Much of the training data is annotated by LLMs. Voting across annotators and checks by judges help reduce shared biases but cannot fully eliminate them. Calibration parameters are fitted on validation data from the training sources, and the resulting probabilities may remain imperfect when the domain changes. We report results from a single seed because repeating training and the full evaluation across 40 benchmarks would require substantial additional computation.

\section*{Ethical Considerations}

When applied to text containing personal information, extraction systems can make sensitive details easier to identify and aggregate. Such applications should respect privacy and the conditions under which the data were obtained. LLM annotations may also retain shared biases despite our annotation checks. Predictions and confidence estimates should be reviewed before they are used in decisions that affect individuals.

\section*{Acknowledgments}

We used LLMs to assist with drafting and restructuring parts of the main text and appendices. The authors reviewed and revised the resulting material and take responsibility for the methods, experimental results, claims, and references in the final manuscript.

% \section*{Acknowledgments}

% Bibliography entries for the entire Anthology, followed by custom entries
%\bibliography{custom,anthology-overleaf-1,anthology-overleaf-2}

% keep the result tables in the main text, before the references
\FloatBarrier

% Custom bibliography entries only
\bibliography{custom}

\appendix

\clearpage

\section{Method Details}
\label{sec:appendix_method}

\paragraph{Brier term.}
The implementation supports an optional multi-class Brier term. For a candidate $c$, the target vector $\mathbf{t}_c$ assigns equal probability to the gold classes in $\mathcal{G}_c$ and zero to the remaining classes. The per-candidate loss is
\begin{equation}
\ell_c = -\log \sum_{g \in \mathcal{G}_c} p_c(g) + \frac{\beta}{2} \lVert \mathbf{p}_c - \mathbf{t}_c \rVert_2^2.
\end{equation}
We set $\beta = 0$ for the reported model, so this term is not included in its training loss (Appendix~\ref{sec:appendix_implementation_details}).

\section{Additional Experimental Details}
\label{sec:appendix_experimental}

\subsection{Data Preparation and Sampling}
\label{sec:appendix_data_preparation}

\paragraph{Sources and schemas.}
Human-labelled sources include Few-NERD with fine-grained entity types \citep{ding-etal-2021-nerd}, MASSIVE with slot annotations \citep{fitzgerald-etal-2023-massive}, and the event datasets MAVEN-Arg \citep{wang-etal-2024-maven}, GENEVA \citep{parekh-etal-2023-geneva}, RAMS \citep{ebner-etal-2020-multi}, and PHEE \citep{sun-etal-2022-phee}. FiNERWeb \citep{golde-etal-2026-finerweb} provides LLM-annotated web passages, and Pile-NER \citep{zhou2024universalner} provides LLM annotations of texts sampled from the Pile.

We use Qwen3.8-27B to generate additional texts and annotations. These sources include new annotations of REBEL texts \citep{huguet-cabot-navigli-2021-rebel-relation} and CC-News articles\footnote{\url{https://commoncrawl.org/blog/news-dataset-available}}, as well as synthetic texts generated from combinations of genres, domains, and schemas. Table~\ref{tab:synth_grid} lists the genres and domains, and Figure~\ref{fig:prompts} gives the generation prompts. 
% Genres include news, scientific writing, reviews, queries, social media, and documents. Domains include science, biomedicine, politics, the arts, business, food, and security.

\paragraph{Checking generated annotations.}
For most generated or reannotated sources, Nemotron-3-Nano-Omni-30B-A3B \citep{nvidia2026nemotron3nanoomni} independently annotates each text. When its annotations differ from those of Qwen3.8-27B, gpt-oss-120b \citep{openai2025gptoss120bgptoss20bmodel} resolves the disagreements.

For a subset of synthetic sources, four additional models annotate each text independently of the generator. These are gpt-oss-120b, Nemotron, Phi-4 \citep{abdin2024phi4}, and Mistral-Small-24B \citep{mistral2025small3}. Together with the generator, they provide five annotations. A candidate mention, relation, trigger, or argument is accepted when a majority of the models that provide an annotation agree on it. This requires three votes when all five models participate. Remaining disagreements are accepted or rejected only when both judges, gpt-oss-120b and Qwen3.8-27B, agree. If a disagreement remains unresolved, or no model provides an annotation for a type, we remove that type from the queried schema for the text. Retained types therefore have a known annotation status.

\paragraph{Duplicate removal and incomplete annotations.}
Within each synthetic source, we remove near-duplicate texts using all-mpnet-base-v2 embeddings and a cosine similarity threshold of 0.9. We also compare training texts with evaluation sentences using word 5-grams. A training text is removed when their overlap covers at least 80\% of the 5-grams in either text. RAMS and GENEVA annotate only some events in each passage. For these sources, event types and argument roles without an annotated occurrence are marked unknown rather than absent.

\paragraph{Sampling.}
We sample with replacement from a pool of 3.1M training instances. With 40,000 updates and a global batch size of 128, the sampling budget is 5.12M draws, including repeated samples. UniMax \citep{chung2023unimax} distributes this budget as evenly as possible across sources. The allocation for each source is capped at four times its number of training instances. The resulting sampling proportions are approximately 35\% entity, 41\% relation, and 23\% event examples.

\subsection{Evaluation Datasets and Scoring}
\label{sec:appendix_evaluation_details}

\paragraph{Entity extraction.}
The NER evaluation covers the five CrossNER domains of AI, literature, music, politics, and science \citep{liu2021crossner}, as well as MIT Movie and MIT Restaurant \citep{liu2013asgard}. It also includes SciER's in-domain and out-of-domain test sets \citep{zhang-etal-2024-scier}, GENTLE \citep{aoyama-etal-2023-gentle}, BioRED \citep{luo-etal-2022-biored}, and MeasEval \citep{harper-etal-2021-semeval}. Personal information extraction is evaluated on ai4privacy \citep{ai4privacy_open_pii_masking_500k} and the legal and medical subsets of spy \citep{savkin-etal-2025-spy}. Table~\ref{tab:datasets} lists the domains, type counts, and evaluation sizes. Each input is queried with the dataset's full entity schema. A queried type with no gold mention is treated as absent for scoring. We split spy documents into chunks of at most 1,200 characters, with boundaries at whitespace outside entity spans.

\paragraph{Relation extraction.}
We evaluate on the six CrossRE domains \citep{bassignana-plank-2022-crossre}, using the relation definitions from the paper as descriptions. Other RE datasets are SciER, BioRED, MeasEval, and Re-TACRED \citep{stoica2021re}. We also represent four ASTE datasets \citep{xu-etal-2020-position} as directed relations from aspect spans to opinion spans, with sentiment as the relation type.

Additional tasks are converted to span-pair extraction. In GAP \citep{webster-etal-2018-mind}, a relation links the designated pronoun to a coreferent person name. In LitBank \citep{bamman-etal-2020-annotated}, each mention is linked to the nearest preceding mention of the same entity within the input. We use all LitBank documents for evaluation. In MAVEN-ERE \citep{wang-etal-2022-maven}, relation candidates are event and time-expression mentions. Temporal, causal, and subevent links use representative mentions of each event within the input. Event coreference links later mentions to the first mention of the same event.

We report only the gold-entity setting for BioRED, GAP, LitBank, MAVEN-ERE, and Re-TACRED. Their converted evaluation inputs use representative concept mentions, designated mentions, or supplied event and coreference mentions. End-to-end extraction is not evaluated under these conversions. Long inputs are divided into chunks outside annotated spans, and links whose endpoints fall in different chunks are excluded for every model. The converted LitBank and MAVEN-ERE scores therefore measure span-pair extraction within an input rather than the original document-level task.

\paragraph{Event extraction.}
We evaluate on CASIE \citep{satyapanich2020casie}, MLEE \citep{pyysalo-etal-2012-mlee}, M2E2 \citep{li-etal-2020-cross}, MAILEX \citep{srivastava-etal-2023-mailex}, SENTiVENT \citep{jacobs2022sentivent}, and SPEED++ \citep{parekh-etal-2024-speed}. For SPEED++, we use the released English files \texttt{en-mpox.json} and \texttt{en-zika.json} and retain mpox and zika as their benchmark labels. We report trigger classification and argument classification with both gold and predicted triggers.

\paragraph{Scoring.}
We compute exact-match micro F1 for each benchmark and give each benchmark equal weight in averages. An entity prediction must match both the character offsets and the type of a gold mention. A relation prediction must match the two spans and the relation type. Directed relations must also match the direction. BioRED relations are symmetric, so reversed endpoints are treated as equivalent. An event argument must match the trigger span and event type, the argument span, and the argument role.

We also measure the false-positive rate on absent types and the miss rate. The former is the fraction of type queries with no gold occurrence for which the model returns at least one prediction. The latter is the fraction of gold mentions without an exact matching prediction.

\subsection{Evaluation Adaptations}
\label{sec:appendix_baseline_details}

\paragraph{Input formats.}
GNER \citep{ding2024rethinking} was trained on Pile-NER \citep{zhou2024universalner}. Following its input format, we split texts into words and separate punctuation into individual tokens. For GoLLIE \citep{sainz2024gollie}, each type is defined as a Python class with its description in the docstring. We leave docstrings empty when evaluating without descriptions and provide no examples in either setting. GLiNER2 \citep{zaratiana2025gliner2} is evaluated on events through its structured extraction interface.

\paragraph{Relations with gold entities.}
\textsc{CalOn} and GLiREL receive gold entity spans directly. For the other baselines, we match the endpoints of their end-to-end relation predictions to gold entities. We first use an exact span match. If none exists, we use the nearest gold mention with the same text, ignoring case. Relations with an unmatched endpoint are discarded. These adapted scores still depend on the baselines identifying the mentions and do not isolate relation prediction in the same way as supplying gold entities directly.

\paragraph{Arguments with gold triggers.}
For GLiNER2 and GoLLIE, we align predicted events with gold triggers of the same event type and retain the arguments of matched events. A match is accepted when the trigger spans are identical, the predicted span contains the gold span, or their trigger texts are identical. InstructUIE predicts arguments for an event type rather than for an individual trigger. We provide the gold event types and assign each predicted argument to the nearest gold trigger of that type.

\paragraph{GLiREL.}
We split texts into words and punctuation, with additional token boundaries at the start and end of every gold entity. GLiREL receives these entities and the relation type names, with hyphens and underscores replaced by spaces. It scores every ordered pair of gold entities, and we retain the highest-scoring type when its score passes the extraction threshold. Predictions assigned the \textit{no relation} label are not returned as relations.

We select the threshold on the REBEL calibration split from $\{0.05, 0.1, 0.2, \ldots, 0.9\}$. On the same split, we compare queries with and without an additional \textit{no relation} label. We retain this label because it gives higher F1, 18.1 compared with 17.2, and select a threshold of 0.05. The evaluated configuration therefore includes both a threshold and a label for rejecting pairs.

We separately measure relation typing when only pairs with a gold relation are supplied. Across the 18 RE benchmarks, GLiREL's mean typing accuracy is 40.0\%, compared with 79.0\% for \textsc{CalOn} when taking its most probable non-null type. With all ordered pairs of gold entities as candidates, their average F1 scores are 2.6 and 49.8. These metrics answer different questions. The first measures typing on positive pairs, while the second also requires rejecting unrelated pairs. GLiREL's low F1 in the latter setting is associated with low precision. Its 100\% typing accuracy on GAP and LitBank is expected because each has only one relation type.

\paragraph{Training coverage.}
The released training configuration associated with the evaluated InstructUIE checkpoint includes MIT Movie, MIT Restaurant, and CASIE\footnote{\url{https://github.com/BeyonderXX/InstructUIE/blob/master/configs/multi_task_configs/train_tasks.json}}. GoLLIE's training includes TACRED and ACE05 \citep{sainz2024gollie}. Re-TACRED revises labels for TACRED sentences \citep{stoica2021re}, while M2E2 uses the ACE event types. These are different forms of prior exposure. The former involves shared source sentences, whereas the latter involves shared event types. We mark the relevant MIT, CASIE, and Re-TACRED scores with $^\ddagger$. Such exposure is relevant to interpreting the results, but the comparison does not isolate its effect from other differences between models.

\subsection{Architecture and Training Settings}
\label{sec:appendix_implementation_details}

\paragraph{Architecture and candidate limits.}
We use bfloat16 mixed precision. Candidate spans contain at most $W = 32$ tokens. Input texts contain at most 1,024 tokens, and the combined text and schema contain at most 3,072 tokens. The projection dimension is $A = 256$, and width embeddings have dimension 32. We use $L = 2$ interaction blocks with four attention heads each, a feed-forward dimension of $4A$, GELU activation \citep{hendrycks2023gelu}, and dropout of 0.1. The bilinear scoring dimension is $r = 64$. Relation candidates are formed from at most 24 entity spans per instance.

\paragraph{Optimisation.}
We train with AdamW optimiser \citep{loshchilov2018decoupled}. The learning rate is $4\times10^{-5}$ for the encoder and $2\times10^{-4}$ for the prediction heads. Weight decay is 0.01, and gradients are clipped at 1.0. The schedule uses 10\% warm-up followed by linear decay. Training runs for 40,000 updates on 32 GPUs with a global batch size of 128, or four instances per GPU. Each type description is omitted independently with probability 0.3. We evaluate every 500 updates on validation data from the training sources, using up to 1,000 instances per source. We retain the checkpoint with the highest mean F1 across these sources and use validation loss to break ties.

\paragraph{Loss settings.}
Using the same validation data, we select $\beta = 0$ and a null-class weight of $\omega = 0.1$. The latter is compared with 0.03 in a shorter preliminary run. The relation null-class weight is also 0.1, and the relation loss coefficient is $\lambda = 1$. Both entity and relation decision rewards use $S = 8$ samples, noise scale $\sigma = 1$, and coefficient $\rho = 0.5$. We increase $\rho$ linearly from zero to 0.5 during the first 200 updates.

\paragraph{Calibration settings.}
For the main results, we fit the entity temperature by minimising the multi-class Brier score on calibration data from the training sources. We search 57 log-spaced temperatures from 0.25 to 4. Extraction thresholds are selected to maximise F1 on a grid from 0.02 to 0.98. The reported uncorrected model uses entity temperature $T = 1.219$ and threshold $\tau = 0.88$. Relations use $T = 1$ and threshold $\tau = 0.85$, fitted on REBEL with gold entities. Event triggers reuse the entity parameters, and event arguments reuse the relation parameters. Appendix~\ref{sec:appendix_calibration} reports the additional correction and refitting analysis.

\section{Additional Results}
\label{sec:appendix_results}

\paragraph{Entities.}
\textsc{CalOn} has the highest average F1 on CrossNER, the other-domain group, and personal information extraction (Table~\ref{tab:ner}). GLiFormer remains stronger on CrossNER Music and Science. InstructUIE leads on MIT Restaurant with 80.9 F1, compared with 66.5 for \textsc{CalOn}. This result raises its MIT average above \textsc{CalOn}'s, but its advantage does not extend to the other NER groups. Both MIT datasets appear in the released training configuration associated with the evaluated InstructUIE checkpoint (Appendix~\ref{sec:appendix_baseline_details}).

\paragraph{Relations.}
\textsc{CalOn} leads every reported RE benchmark in both the gold-entity and available end-to-end settings (Table~\ref{tab:re}). This includes conventional relations, ASTE, and pronoun resolution. Its end-to-end CrossRE average is 36.0 F1, compared with 10.6 for the strongest baseline. Scores with gold entities are zero for several baselines on GAP, LitBank, and MAVEN-ERE. However, \textsc{CalOn}'s scores on the latter two also remain low, so its lead does not imply that these tasks are solved. Dashes indicate an unevaluated end-to-end conversion or GLiREL's unsupported end-to-end setting, as described in Appendix~\ref{sec:appendix_evaluation_details}. The GLiNER2 paper reports NER and text-classification benchmarks but no RE or EE benchmark results\footnote{A GLiNER2 maintainer reports that fewer than 5\% of its data contain relation annotations at \url{https://github.com/fastino-ai/GLiNER2/issues/3}. The full training data for the evaluated checkpoint are unavailable to us. We therefore evaluate the released weights rather than reproduce its training run.}.

\paragraph{Events.}
\textsc{CalOn} has the highest average trigger F1 and argument F1 with gold triggers (Table~\ref{tab:ee}). It leads five of the seven datasets on arguments with gold triggers. InstructUIE's CASIE end-to-end argument score of 49.4, compared with 14.9 for \textsc{CalOn}, raises its overall average above \textsc{CalOn}'s. CASIE appears in its released training configuration. Excluding CASIE, \textsc{CalOn} has the highest average for all three EE metrics. Its gold-trigger argument F1 is 34.7 compared with InstructUIE's 28.4, and its end-to-end argument F1 is 10.9 compared with 8.4.

Individual exceptions remain. InstructUIE is stronger on M2E2 arguments with gold triggers and on SPEED++ zika triggers and end-to-end arguments. GoLLIE leads on M2E2 triggers and end-to-end arguments. InstructUIE scores zero on M2E2 triggers and end-to-end arguments and on MAILEX end-to-end arguments. GLiNER2 scores zero on MLEE and MAILEX arguments with gold triggers. These differences show why dataset-level results and training coverage should accompany task averages.

\input{tables/familiarity}

\section{Results without Type Descriptions}
\label{sec:appendix_wo_desc}

Tables~\ref{tab:ner_wo_desc}, \ref{tab:re_wo_desc}, and \ref{tab:ee_wo_desc} repeat the evaluation with type names alone. Models that support descriptions therefore receive the same schema information as models that accept only names. \textsc{CalOn} is trained with each description omitted independently with probability 0.3 (Appendix~\ref{sec:appendix_implementation_details}).

For NER, \textsc{CalOn} retains the highest averages on the other-domain group and personal information extraction. These are 53.7 and 57.2 F1, compared with the strongest baseline averages of 41.5 and 54.2. GLiFormer leads on CrossNER with 67.6 F1, compared with 66.2 for \textsc{CalOn}. For RE, \textsc{CalOn} leads every reported benchmark in both settings. Its gold-entity CrossRE average falls from 45.4 with descriptions to 29.1 without them, indicating that the relation definitions provide useful information beyond the names.

For EE, \textsc{CalOn} retains the highest average trigger F1. Its argument F1 with gold triggers falls from 35.1 to 32.9, slightly below InstructUIE's 33.2. InstructUIE also retains the highest end-to-end argument average. These results show that the effect of descriptions varies across tasks and datasets.

\input{tables/ner_without_description}
\input{tables/re_without_description}
\input{tables/ee_without_description}

\section{Calibration per Benchmark}
\label{sec:appendix_calibration}

Table~\ref{tab:calibration} reports F1 and expected calibration error (ECE) for the original fitted model, null-class correction with threshold refitting, and correction with temperature and threshold refitting. Entity calibration uses validation data from the training sources. Relation calibration uses the REBEL calibration split. The default relation decoder uses $T = 1$, while the temperature-refitting condition is an additional calibration analysis. Event triggers reuse the entity parameters, and event arguments reuse the relation parameters. No event-specific parameters are fitted.

ECE is computed over returned predictions using 15 equal-width confidence bins. The confidence is the probability assigned to the predicted type or argument role. Correctness requires the exact span and type or role matches described in Appendix~\ref{sec:appendix_evaluation_details}. RE is evaluated with gold entities, and event arguments are evaluated with gold triggers. This ECE measures the confidence of returned predictions rather than calibration over all null and non-null candidates. 
Correction and temperature refitting reduce average ECE from 0.33 to 0.23 for entities, from 0.25 to 0.09 for relations, and from 0.51 to 0.39 for event arguments, with little change in F1. Relation ECE decreases on 17 of the 18 benchmarks. CrossRE News is the exception, with an initial ECE of 0.08. On the REBEL calibration split, ECE is 0.13 before correction, 0.01 after correction and threshold refitting, and 0.03 after additional temperature refitting. Thus, temperature refitting does not improve every individual result.

Argument ECE decreases on all seven event benchmarks. MLEE improves from 0.29 to 0.11, but ECE remains at least 0.30 on the other six datasets. The multi-class Brier score over all relation candidates changes little, with averages of 0.157, 0.162, and 0.158 across the three settings. Since most candidate pairs are null, this metric emphasises a different set of decisions from ECE over returned relations.

\input{tables/calibration}

\section{Type Familiarity}
\label{sec:appendix_familiarity}

\citet{golde2025familiarity} study zero-shot NER in terms of type-name similarity and training frequency. All entity type names in our 15 NER benchmarks occur verbatim in \textsc{CalOn}'s training data. These benchmarks therefore test transfer across domains and annotation guidelines rather than to unseen entity type names.

We examine whether performance depends on how frequently a type name is annotated in training. The 150 evaluated entity types, counted separately for each benchmark, are divided into thirds using their mention counts in \textsc{CalOn}'s training data (Table~\ref{tab:familiarity}). We report macro F1 over types in each group and the Spearman correlation between training frequency and per-type F1.

\textsc{CalOn} has the highest macro F1 in all three groups. Its correlation with training frequency is small, $\rho = 0.04$, and its largest margin is in the rarest group, with 58.8 F1 compared with GLiNER2's 50.6. The generative baselines have positive correlations from 0.10 to 0.21 under the same grouping. However, these frequencies come from \textsc{CalOn}'s training data and do not measure the baselines' own training exposure. The analysis shows that \textsc{CalOn}'s gains are not confined to its most frequent type names. It does not isolate the effects of the training data and model architecture.

\section{Inference Cost}
\label{sec:appendix_efficiency}

Table~\ref{tab:efficiency} compares inference on one GH200 GPU, with one model loaded at a time. All models use PyTorch and Transformers with bf16 weights. We time the 1,953 MIT Movie test sentences with all 12 entity types and exclude model loading. Generative models use greedy decoding and batches of 32. Their MIT Movie F1 is unchanged from the main evaluation.

\textsc{CalOn} processes 120 sentences per second. This is about five to eight times the throughput of GLiNER2, GLiFormer, and InstructUIE, 11 times that of GoLLIE, and 25 times that of GNER. Its weight memory is 1.2 GB, compared with 13.5 GB for the 7B models. Peak GPU memory is 8.0 GB, higher than GLiNER2 and GLiFormer but lower than the generative baselines. Parameter count alone therefore does not capture the memory cost of inference. 
These timings depend on the inference framework. With vLLM, GNER and GoLLIE reach 112 and 238 sentences per second, respectively. This setting uses continuous batching and shared-prefix caching and reserves most of the GPU memory. GoLLIE is faster than \textsc{CalOn} under this separate setup, so the PyTorch comparison does not establish a speed advantage across serving frameworks. Our \textsc{CalOn} evaluation also includes calibration-statistic computation and is not optimised solely for throughput. The table reports the cost of this evaluation implementation.

\section{Data Scaling}
\label{sec:appendix_scaling}

We train on nested random subsets of 100K, 250K, 500K, 1M, and 2M instances and compare them with the full 3.1M-instance model. Each subset covers all 84 sources and preserves their proportions in the full sampling mixture. Smaller subsets are contained in larger ones. Architecture, objective, batch size, and checkpoint selection follow the full model. Training updates scale with dataset size, from 1,282 for 100K instances to 40,000 for the full data, so each run makes approximately the same number of passes over its data. Evaluation intervals are also scaled to give 80 evaluations per run. Each model is trained with one seed.

\begin{figure}[t]
\centering
\includegraphics[width=\columnwidth]{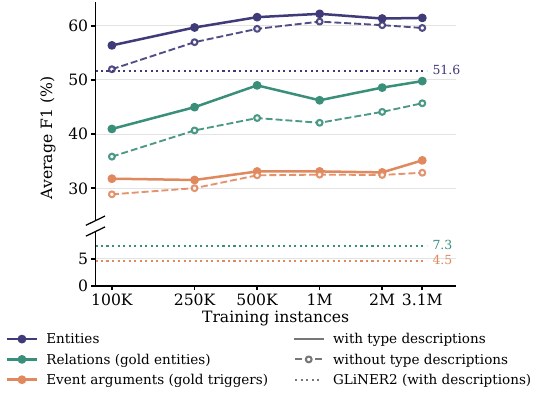}
\caption{Average micro F1 as training data increase. Solid lines use type descriptions, dashed lines use type names alone, and dotted lines show GLiNER2 with descriptions. Averages cover the 15 NER, 18 RE with gold entities, and 7 EE with gold triggers benchmarks.}
\label{fig:scaling}
\end{figure}

Figure~\ref{fig:scaling} reports average NER F1, RE F1 with gold entities, and event-argument F1 with gold triggers. With descriptions, entity and relation averages improve by approximately five and eight points from 100K to 500K instances. Gains after 500K are smaller and not monotonic. Event-argument F1 changes less across the smaller subsets and improves by about two points with the full data. Results with type names alone show a similar pattern. Since training updates increase with dataset size, the experiment varies data and training compute together. The single-seed results do not establish whether the smaller differences are statistically reliable.

\section{Models}
\label{sec:appendix_models}

Table~\ref{tab:models} identifies the checkpoints used for the \textsc{CalOn} backbone, baselines, data generation and annotation, and text-embedding analysis. It also lists their model links and reported licenses.

\input{tables/models}

\section{Evaluation Datasets}
\label{sec:appendix_datasets}

Table~\ref{tab:datasets} lists the 40 evaluation benchmarks with their tasks, domains, genres, type counts, and evaluation sizes. Example counts refer to the inputs after conversion to our evaluation format, which may be sentences, passages, or document chunks. The task conversions and scoring rules are described in Appendix~\ref{sec:appendix_evaluation_details}.

\input{tables/datasets}

\input{tables/train_datasets}

\input{tables/synth_grid}

\section{Training Data}
\label{sec:appendix_data}

Table~\ref{tab:train_datasets} lists training sources that use real texts and identifies whether their annotations come from humans, released LLM annotations, or our annotation pipeline. Counts refer to the training instances used in this work rather than the sizes of the original datasets. 
Table~\ref{tab:synth_grid} lists the genres and domains used for synthetic generation. Each request combines one genre with one domain. The prompts in Figure~\ref{fig:prompts} ask the generator to write a text, define an extraction schema, annotate the text, and list additional types that do not occur. These absent types provide explicit negative supervision after annotation checking.

\begin{figure*}[t]
\centering
\includegraphics[width=\textwidth]{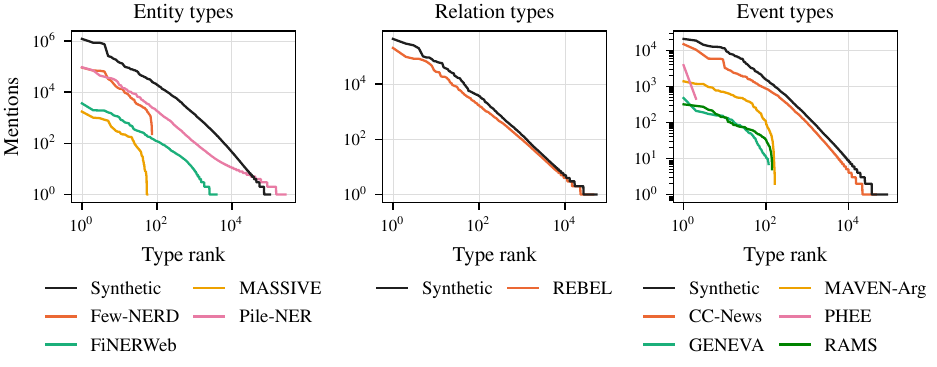}
\caption{Training type frequencies. For each source, types are ranked by mention count. Both axes use logarithmic scales. Longer curves indicate more type names, and flatter curves indicate a more even distribution of mentions. The synthetic curve pools all synthetic sources, while the other curves show individual datasets. Counts use the complete training files before sampling.}
\label{fig:long_tail}
\end{figure*}

\input{tables/prompts}

\begin{figure*}[tp]
\centering
\includegraphics[width=\columnwidth]{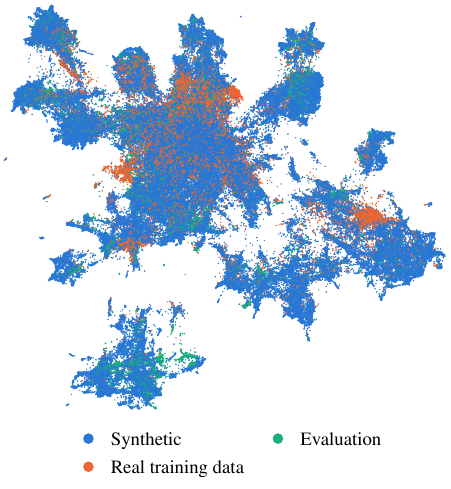}
\caption{UMAP \citep{mcinnes2018umap-software} projection of sentence embeddings sampled from synthetic training sources, real-text training sources, and evaluation datasets.}
\label{fig:umap}

\vspace{2ex}
\includegraphics[width=\textwidth]{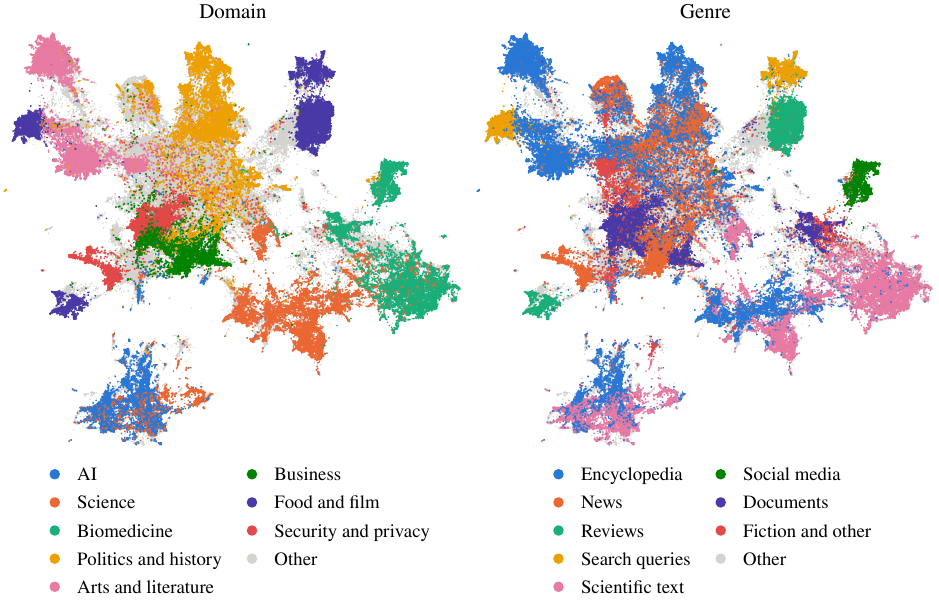}
\caption{The projection in Figure~\ref{fig:umap} coloured by domain on the left and genre on the right. Texts outside the listed categories are shown in grey.}
\label{fig:umap_domain}
\end{figure*}

\paragraph{Type distribution.}
Figure~\ref{fig:long_tail} shows type-name frequencies before training-source sampling. Human-labelled sources have fixed inventories, with 74 entity types in Few-NERD, 55 slots in MASSIVE, and between two and 162 event types in the event datasets. Pile-NER and the re-annotated REBEL texts have broader inventories with many infrequent type names. The pooled synthetic sources contain approximately 105K entity type names, 54K relation type names, and 85K event type names. These counts describe names used in the annotations and do not assume that every distinct name denotes a distinct semantic category.

\paragraph{Domains and genres.}
We sample 3,000 texts per training source and up to 500 per evaluation set. Texts are encoded with the all-mpnet-base-v2 checkpoint using the SentenceTransformers framework \citep{reimers-gurevych-2019-sentence} and projected into two dimensions with UMAP. Synthetic texts occupy a broader range of regions than the real-text training sources in this projection. Several evaluation regions also contain synthetic texts, including regions associated with reviews, search queries, social media, and personal information. Figure~\ref{fig:umap_domain} shows how these regions relate to the assigned domains and genres. This is a qualitative view of the sampled embeddings. Overlap in the projection does not establish equivalent data distributions or explain the performance gains on its own.

\end{document}

%% file: tables/ner_with_description.tex
\begin{table*}[t]
\centering
\small
\setlength{\tabcolsep}{3.5pt}
\begin{adjustbox}{max width=\textwidth}
\begin{tabular}{lrrrrrr}
\toprule
Dataset & \textsc{CalOn} & GLiNER2 & GLiFormer$^\dagger$ & GNER$^\dagger$ & InstructUIE$^\dagger$ & GoLLIE \\
\midrule
\multicolumn{7}{l}{\textit{CrossNER}} \\
AI & \textbf{66.1} & 56.1 & \underline{59.1} & 56.3 & 48.1 & 49.2 \\
Literature & \textbf{63.5} & 62.5 & \underline{62.8} & \textbf{63.5} & 44.2 & 55.7 \\
Music & \underline{72.0} & 67.7 & \textbf{76.2} & 71.1 & 58.8 & 62.8 \\
Politics & \textbf{73.0} & 65.2 & \underline{70.6} & 63.8 & 45.7 & 47.5 \\
Science & \underline{64.4} & 60.5 & \textbf{69.0} & 60.8 & 45.9 & 48.5 \\
\cmidrule(lr){1-7}
Average & \textbf{67.8} & 62.4 & \underline{67.6} & 63.1 & 48.5 & 52.7 \\
\midrule
\multicolumn{7}{l}{\textit{MIT}} \\
Movie & \textbf{58.1} & 50.6 & \underline{57.5} & 54.5 & 52.8$^\ddagger$ & 52.3 \\
Restaurant & \underline{66.5} & 55.3 & 61.7 & 46.6 & \textbf{80.9}$^\ddagger$ & 24.2 \\
\cmidrule(lr){1-7}
Average & \underline{62.3} & 53.0 & 59.6 & 50.6 & \textbf{66.9}$^\ddagger$ & 38.2 \\
\midrule
\multicolumn{7}{l}{\textit{Other domains}} \\
SciER & \textbf{70.9} & 51.6 & 51.9 & 46.1 & 51.4 & \underline{54.6} \\
SciER (OOD) & \textbf{67.2} & \underline{49.5} & 48.4 & 41.7 & 45.8 & 32.8 \\
GENTLE & \textbf{42.0} & 24.1 & \underline{27.2} & 11.3 & 15.5 & 16.0 \\
BioRED & \textbf{72.1} & 57.3 & 57.5 & 43.4 & 33.7 & \underline{60.4} \\
MeasEval & \textbf{30.6} & \underline{22.7} & 22.4 & 3.9 & 7.4 & 17.7 \\
\cmidrule(lr){1-7}
Average & \textbf{56.6} & 41.1 & \underline{41.5} & 29.3 & 30.8 & 36.3 \\
\midrule
\multicolumn{7}{l}{\textit{Personal information}} \\
ai4privacy & \textbf{68.8} & 55.0 & \underline{56.1} & 36.6 & 34.8 & 51.6 \\
spy legal & \textbf{53.2} & 47.4 & \underline{52.8} & 20.1 & 50.3 & 44.8 \\
spy medical & \textbf{53.5} & 48.2 & \textbf{53.5} & 21.7 & \underline{49.4} & 46.1 \\
\cmidrule(lr){1-7}
Average & \textbf{58.5} & 50.2 & \underline{54.2} & 26.1 & 44.8 & 47.5 \\
\bottomrule
\end{tabular}
\end{adjustbox}
\caption{Zero-shot NER, micro F1 in \%. For each sentence, every model receives the full type set of its dataset (e.g., all 14 types of CrossNER AI, including types that do not occur in the sentence), each with a short description; models marked $^\dagger$ accept only type names and receive the names alone. $^\ddagger$: the model was trained on the training split of this dataset. The best score per row is in \textbf{bold}, and the second-highest distinct score is \underline{underlined}. 
% \textcolor{red}{[CL: can try to highlight the second best with underline]}
% results with type names only for all models are in Table~\ref{tab:ner_wo_desc}.
}
\label{tab:ner}
\end{table*}

%% file: tables/re_with_description.tex
\begin{table*}[t]
\centering
\small
\setlength{\tabcolsep}{3.5pt}
\begin{adjustbox}{max width=\textwidth}
\begin{tabular}{lccccccccccccc}
\toprule
\multirow{2}{*}{Dataset} & \multicolumn{6}{c}{Gold entities} & \multicolumn{6}{c}{End-to-End} \\
\cmidrule(lr){2-7}\cmidrule(lr){8-13}
 & \textsc{CalOn} & GLiNER2 & GLiFormer$^\dagger$ & GLiREL$^\dagger$ & InstructUIE$^\dagger$ & GoLLIE & \textsc{CalOn} & GLiNER2 & GLiFormer$^\dagger$ & GLiREL$^\dagger$ & InstructUIE$^\dagger$ & GoLLIE \\
\midrule
\multicolumn{13}{l}{\textit{CrossRE}} \\
AI & \textbf{33.6} & 7.5 & 7.5 & 3.8 & 8.9 & 6.3 & \textbf{24.6} & 6.4 & 7.0 & -- & 6.5 & 5.2 \\
Literature & \textbf{48.2} & 18.2 & 9.5 & 1.7 & 3.9 & 20.5 & \textbf{35.0} & 15.3 & 9.1 & -- & 3.2 & 15.5 \\
Music & \textbf{51.6} & 9.5 & 7.9 & 2.4 & 3.9 & 13.6 & \textbf{40.1} & 8.3 & 7.6 & -- & 3.4 & 10.3 \\
News & \textbf{61.9} & 22.1 & 22.2 & 6.7 & 4.8 & 1.5 & \textbf{58.4} & 20.0 & 22.0 & -- & 3.8 & 1.3 \\
Politics & \textbf{40.4} & 5.2 & 7.4 & 3.2 & 2.3 & 6.8 & \textbf{32.6} & 4.6 & 7.3 & -- & 2.0 & 5.0 \\
Science & \textbf{36.8} & 7.3 & 11.3 & 3.8 & 5.0 & 11.2 & \textbf{25.5} & 5.8 & 10.4 & -- & 3.8 & 9.3 \\
\cmidrule(lr){1-13}
Average & \textbf{45.4} & 11.6 & 11.0 & 3.6 & 4.8 & 10.0 & \textbf{36.0} & 10.1 & 10.6 & -- & 3.8 & 7.8 \\
\midrule
\multicolumn{13}{l}{\textit{ASTE}} \\
14lap & \textbf{75.6} & 2.5 & 0.7 & 0.0 & 8.9 & 0.0 & \textbf{43.5} & 2.3 & 0.7 & -- & 6.5 & 0.0 \\
14res & \textbf{86.3} & 5.0 & 0.8 & 0.0 & 20.9 & 0.0 & \textbf{61.2} & 4.8 & 0.8 & -- & 17.8 & 0.0 \\
15res & \textbf{83.9} & 5.1 & 1.2 & 0.0 & 20.0 & 0.0 & \textbf{54.5} & 4.8 & 1.2 & -- & 15.9 & 0.0 \\
16res & \textbf{86.3} & 7.0 & 1.9 & 0.0 & 18.7 & 0.0 & \textbf{60.5} & 6.6 & 1.9 & -- & 14.9 & 0.0 \\
\cmidrule(lr){1-13}
Average & \textbf{83.0} & 4.9 & 1.2 & 0.0 & 17.1 & 0.0 & \textbf{55.0} & 4.6 & 1.2 & -- & 13.7 & 0.0 \\
\midrule
\multicolumn{13}{l}{\textit{Other tasks}} \\
SciER & \textbf{43.4} & 10.6 & 24.1 & 1.4 & 17.1 & 1.5 & \textbf{27.0} & 8.0 & 18.7 & -- & 10.1 & 1.4 \\
SciER (OOD) & \textbf{38.9} & 10.3 & 16.1 & 4.5 & 20.7 & 0.7 & \textbf{22.0} & 7.6 & 10.9 & -- & 10.3 & 0.7 \\
BioRED & \textbf{17.7} & 0.8 & 0.0 & 1.1 & 2.3 & 1.2 & -- & -- & -- & -- & -- & -- \\
GAP & \textbf{76.3} & 0.0 & 0.0 & 1.4 & 0.0 & 3.9 & -- & -- & -- & -- & -- & -- \\
LitBank & \textbf{16.8} & 0.0 & 0.0 & 0.1 & 0.0 & 0.0 & -- & -- & -- & -- & -- & -- \\
MAVEN-ERE & \textbf{2.0} & 0.0 & 0.0 & 0.0 & 0.0 & 0.0 & -- & -- & -- & -- & -- & -- \\
MeasEval & \textbf{57.9} & 2.1 & 8.9 & 0.0 & 1.0 & 0.8 & \textbf{16.6} & 2.0 & 6.2 & -- & 1.0 & 0.6 \\
Re-TACRED & \textbf{38.4} & 18.3 & 13.8 & 16.0 & 11.4 & 20.5$^\ddagger$ & -- & -- & -- & -- & -- & -- \\
\cmidrule(lr){1-13}
Average & \textbf{36.4} & 5.3 & 7.9 & 3.1 & 6.6 & 3.6$^\ddagger$ & \textbf{21.9} & 5.8 & 11.9 & -- & 7.1 & 0.9 \\
\bottomrule
\end{tabular}
\end{adjustbox}
\caption{RE micro F1 (\%). \emph{Gold entities} uses supplied gold spans for \textsc{CalOn} and GLiREL and adapted baseline predictions matched to gold entities (Appendix~\ref{sec:appendix_baseline_details}). \emph{End-to-end} predicts entities and relations. Dashes mark unevaluated end-to-end conversions or GLiREL's unsupported end-to-end setting. $^\dagger$ marks type names alone. $^\ddagger$ marks GoLLIE, which was trained on TACRED sentences reused by Re-TACRED. Best scores are bold. 
% Results without descriptions are in Table~\ref{tab:re_wo_desc}.
}
\label{tab:re}
\end{table*}

%% file: tables/ee_with_description.tex
\begin{table*}[t]
\centering
\small
\setlength{\tabcolsep}{3.5pt}
\begin{adjustbox}{max width=\textwidth}
\begin{tabular}{lrrrrrrrrrrrr}
\toprule
\multirow{2}{*}{Dataset} & \multicolumn{4}{c}{Arg.\ (gold triggers)} & \multicolumn{4}{c}{Trigger (end-to-end)} & \multicolumn{4}{c}{Arg.\ (end-to-end)} \\
\cmidrule(lr){2-5}\cmidrule(lr){6-9}\cmidrule(lr){10-13}
 & \textsc{CalOn} & GLiNER2 & InstructUIE$^\dagger$ & GoLLIE & \textsc{CalOn} & GLiNER2 & InstructUIE$^\dagger$ & GoLLIE & \textsc{CalOn} & GLiNER2 & InstructUIE$^\dagger$ & GoLLIE \\
\midrule
CASIE & 37.9 & 3.4 & \textbf{62.6}$^\ddagger$ & 3.9 & 33.8 & 7.5 & \textbf{67.6}$^\ddagger$ & 10.1 & 14.9 & 3.1 & \textbf{49.4}$^\ddagger$ & 3.2 \\
MLEE & \textbf{11.8} & 0.0 & 3.7 & 0.2 & \textbf{52.0} & 0.2 & 3.7 & 3.2 & \textbf{4.5} & 0.0 & 0.6 & 0.2 \\
M2E2 & 30.9 & 15.5 & \textbf{50.2} & 33.6 & 9.2 & 21.4 & 0.0 & \textbf{38.3} & 3.6 & 8.1 & 0.0 & \textbf{18.3} \\
MAILEX & \textbf{23.9} & 0.0 & 4.7 & 0.0 & \textbf{9.3} & 0.0 & 0.5 & 0.4 & \textbf{2.2} & 0.0 & 0.0 & 0.0 \\
SENTiVENT & \textbf{37.7} & 0.3 & 28.5 & 3.7 & \textbf{14.7} & 0.4 & 11.8 & 8.8 & \textbf{7.6} & 0.2 & 5.0 & 3.1 \\
SPEED++ mpox & \textbf{46.9} & 1.3 & 37.5 & 10.5 & \textbf{41.0} & 2.8 & 35.7 & 13.2 & \textbf{22.1} & 1.1 & 18.3 & 7.6 \\
SPEED++ zika & \textbf{56.8} & 11.2 & 45.5 & 18.2 & 42.0 & 15.0 & \textbf{49.4} & 25.5 & 25.5 & 10.3 & \textbf{26.5} & 14.6 \\
\cmidrule(lr){1-13}
Average & \textbf{35.1} & 4.5 & 33.2$^\ddagger$ & 10.0 & \textbf{28.9} & 6.8 & 24.1$^\ddagger$ & 14.2 & 11.5 & 3.3 & \textbf{14.3}$^\ddagger$ & 6.7 \\
\bottomrule
\end{tabular}
\end{adjustbox}
\caption{Event extraction, micro F1 in \%. \emph{Arg.\ (gold triggers)}: argument and role classification given the gold triggers; \emph{Trigger} and \emph{Arg.\ (end to end)}: trigger classification, and argument classification on the predicted triggers. Models marked $^\dagger$ accept only type names and receive the names alone. $^\ddagger$: the model was trained on the training split of this dataset. Best score per row in \textbf{bold}. 
% results with type names only (i.e., without description) are in Table~\ref{tab:ee_wo_desc}.
}
\label{tab:ee}
\end{table*}

%% file: tables/efficiency.tex
\begin{table}[t]
\centering
\small
\setlength{\tabcolsep}{3pt}
\begin{adjustbox}{max width=\columnwidth}
\begin{tabular}{lrrrrr}
\toprule
Model & Params & Weights (GB) & Peak (GB) & Sent./s & NER F1 \\
\midrule
\textsc{CalOn} & 0.6B & 1.2 & 8.0 & 120 & \textbf{61.5} \\
GLiNER2 & 0.5B & 1.0 & 2.9 & 22 & 51.6 \\
GLiFormer & 0.6B & 1.2 & 3.2 & 18 & 55.1 \\
GNER & 6.7B & 13.5 & 28.2 & 4.9 & 42.8 \\
InstructUIE & 11.4B & 22.8 & 28.0 & 16 & 44.3 \\
GoLLIE & 6.7B & 13.5 & 25.1 & 10.8 & 44.3 \\
\bottomrule
\end{tabular}
\end{adjustbox}
% \caption{Inference cost on one GH200 GPU, one model at a time, all in PyTorch / transformers (generative models: greedy decoding, batches of 32). \emph{Weights}: parameters in bf16; \emph{Peak}: highest GPU memory in use; \emph{Sent./s}: MIT Movie test sentences (1,953, all 12 types) per second, excluding model loading; \emph{NER F1}: average over the 15 NER benchmarks of Table~\ref{tab:ner}.}
\caption{Inference cost on a GH200 GPU using PyTorch and Transformers with bf16 weights. Throughput is measured on MIT Movie with all 12 types, excluding model loading. Generative models use greedy decoding with batch size 32. NER F1 is averaged over the 15 benchmarks in Table~\ref{tab:ner}.}
\label{tab:efficiency}
\end{table}

%% file: tables/familiarity.tex
\begin{table}[t]
\centering
\small
\setlength{\tabcolsep}{4pt}
\begin{tabular}{l*{3}{>{\centering\arraybackslash}p{1.35cm}}c}
\toprule
\multirow{2}{*}{Model} & \multicolumn{3}{c}{Training mentions of the name} & \multirow{2}{*}{$\rho$} \\
\cmidrule(lr){2-4}
 & Rare & Middle & Frequent & \\
\midrule
\textsc{CalOn} & \textbf{58.8} & \textbf{54.8} & \textbf{61.3} & 0.04 \\
GLiNER2 & 50.6 & 46.7 & 52.0 & 0.01 \\
GLiFormer & 50.0 & 50.6 & 55.2 & 0.09 \\
GNER & 36.1 & 42.7 & 48.7 & 0.21 \\
InstructUIE & 38.7 & 39.5 & 44.2 & 0.10 \\
GoLLIE & 37.9 & 39.5 & 48.4 & 0.16 \\
\bottomrule
\end{tabular}
\caption{Macro F1 (\%) over the 150 entity types in the 15 NER benchmarks, counting types separately for each benchmark. Types are grouped into thirds by their mention counts in \textsc{CalOn}'s training data. Rare types have at most 13,400 mentions, and frequent types have more than 77,924. $\rho$ is the Spearman correlation between these counts and per-type F1. Evaluation settings follow Table~\ref{tab:ner}.}
\label{tab:familiarity}
\end{table}

%% file: tables/ner_without_description.tex
\begin{table*}[t]
\centering
\small
\setlength{\tabcolsep}{3.5pt}
\begin{adjustbox}{max width=\textwidth}
\begin{tabular}{lrrrrrr}
\toprule
Dataset & \textsc{CalOn} & GLiNER2 & GLiFormer & GNER & InstructUIE & GoLLIE \\
\midrule
\multicolumn{7}{l}{\textit{CrossNER}} \\
AI & \textbf{63.3} & 51.0 & \underline{59.1} & 56.3 & 48.1 & 37.5 \\
Literature & \underline{63.2} & 57.9 & 62.8 & \textbf{63.5} & 44.2 & 51.0 \\
Music & \underline{71.4} & 64.0 & \textbf{76.2} & 71.1 & 58.8 & 56.3 \\
Politics & \textbf{72.5} & 66.8 & \underline{70.6} & 63.8 & 45.7 & 38.2 \\
Science & \underline{60.8} & 53.6 & \textbf{69.0} & \underline{60.8} & 45.9 & 42.0 \\
\cmidrule(lr){1-7}
Average & \underline{66.2} & 58.7 & \textbf{67.6} & 63.1 & 48.5 & 45.0 \\
\midrule
\multicolumn{7}{l}{\textit{MIT}} \\
Movie & \underline{56.8} & 45.1 & \textbf{57.5} & 54.5 & 52.8$^\ddagger$ & 49.5 \\
Restaurant & \underline{65.8} & 45.0 & 61.7 & 46.6 & \textbf{80.9}$^\ddagger$ & 16.8 \\
\cmidrule(lr){1-7}
Average & \underline{61.3} & 45.0 & 59.6 & 50.6 & \textbf{66.9}$^\ddagger$ & 33.1 \\
\midrule
\multicolumn{7}{l}{\textit{Other domains}} \\
SciER & \textbf{67.8} & 37.6 & \underline{51.9} & 46.1 & 51.4 & 17.6 \\
SciER (OOD) & \textbf{60.9} & 37.9 & \underline{48.4} & 41.7 & 45.8 & 14.4 \\
GENTLE & \textbf{38.1} & 27.1 & \underline{27.2} & 11.3 & 15.5 & 5.4 \\
BioRED & \textbf{70.9} & 47.8 & \underline{57.5} & 43.4 & 33.7 & 57.2 \\
MeasEval & \textbf{31.1} & 13.8 & \underline{22.4} & 3.9 & 7.4 & 1.5 \\
\cmidrule(lr){1-7}
Average & \textbf{53.7} & 32.9 & \underline{41.5} & 29.3 & 30.8 & 19.2 \\
\midrule
\multicolumn{7}{l}{\textit{Personal information}} \\
ai4privacy & \textbf{66.0} & \underline{56.9} & 56.1 & 36.6 & 34.8 & 46.8 \\
spy legal & \underline{52.6} & 48.2 & \textbf{52.8} & 20.1 & 50.3 & 42.4 \\
spy medical & \underline{52.8} & 48.7 & \textbf{53.5} & 21.7 & 49.4 & 44.8 \\
\cmidrule(lr){1-7}
Average & \textbf{57.2} & 51.3 & \underline{54.2} & 26.1 & 44.8 & 44.7 \\
\bottomrule
\end{tabular}
\end{adjustbox}
\caption{NER with type names alone, reported as micro F1 (\%). Evaluation settings follow Table~\ref{tab:ner}. $^\ddagger$ marks scores for the evaluated InstructUIE checkpoint whose released training configuration includes the corresponding dataset. Best scores are \textbf{bold}; second-highest distinct scores are \underline{underlined}.}
\label{tab:ner_wo_desc}
\end{table*}

%% file: tables/re_without_description.tex
\begin{table*}[t]
\centering
\small
\setlength{\tabcolsep}{3.5pt}
\begin{adjustbox}{max width=\textwidth}
\begin{tabular}{lrrrrrrrrrrrr}
\toprule
\multirow{2}{*}{Dataset} & \multicolumn{6}{c}{Gold entities} & \multicolumn{6}{c}{End-to-end} \\
\cmidrule(lr){2-7}\cmidrule(lr){8-13}
 & \textsc{CalOn} & GLiNER2 & GLiFormer & GLiREL & InstructUIE & GoLLIE & \textsc{CalOn} & GLiNER2 & GLiFormer & GLiREL & InstructUIE & GoLLIE \\
\midrule
\multicolumn{13}{l}{\textit{CrossRE}} \\
AI & \textbf{28.4} & 6.5 & 7.5 & 3.8 & 8.9 & 0.0 & \textbf{20.0} & 4.3 & 7.0 & -- & 6.5 & 0.0 \\
Literature & \textbf{23.0} & 5.0 & 9.5 & 1.7 & 3.9 & 0.6 & \textbf{17.6} & 3.5 & 9.1 & -- & 3.2 & 0.6 \\
Music & \textbf{24.0} & 2.2 & 7.9 & 2.4 & 3.9 & 0.0 & \textbf{18.9} & 1.7 & 7.6 & -- & 3.4 & 0.0 \\
News & \textbf{41.4} & 10.4 & 22.2 & 6.7 & 4.8 & 0.0 & \textbf{36.7} & 8.1 & 22.0 & -- & 3.8 & 0.0 \\
Politics & \textbf{29.5} & 2.8 & 7.4 & 3.2 & 2.3 & 0.8 & \textbf{24.9} & 2.1 & 7.3 & -- & 2.0 & 0.8 \\
Science & \textbf{28.6} & 7.0 & 11.3 & 3.8 & 5.0 & 0.3 & \textbf{20.3} & 4.9 & 10.4 & -- & 3.8 & 0.3 \\
\cmidrule(lr){1-13}
Average & \textbf{29.1} & 5.6 & 11.0 & 3.6 & 4.8 & 0.3 & \textbf{23.0} & 4.1 & 10.6 & -- & 3.8 & 0.3 \\
\midrule
\multicolumn{13}{l}{\textit{ASTE}} \\
14lap & \textbf{70.2} & 23.6 & 0.7 & 0.0 & 8.9 & 0.0 & \textbf{24.9} & 20.8 & 0.7 & -- & 6.5 & 0.0 \\
14res & \textbf{83.0} & 38.4 & 0.8 & 0.0 & 20.9 & 0.0 & \textbf{45.0} & 34.6 & 0.8 & -- & 17.8 & 0.0 \\
15res & \textbf{81.2} & 44.0 & 1.2 & 0.0 & 20.0 & 0.0 & \textbf{40.6} & 37.5 & 1.2 & -- & 15.9 & 0.0 \\
16res & \textbf{85.9} & 41.2 & 1.9 & 0.0 & 18.7 & 0.0 & \textbf{42.3} & 35.6 & 1.9 & -- & 14.9 & 0.0 \\
\cmidrule(lr){1-13}
Average & \textbf{80.1} & 36.8 & 1.2 & 0.0 & 17.1 & 0.0 & \textbf{38.2} & 32.1 & 1.2 & -- & 13.7 & 0.0 \\
\midrule
\multicolumn{13}{l}{\textit{Other tasks}} \\
SciER & \textbf{46.7} & 14.0 & 24.1 & 1.4 & 17.1 & 0.6 & \textbf{28.7} & 8.6 & 18.7 & -- & 10.1 & 0.6 \\
SciER (OOD) & \textbf{45.6} & 13.4 & 16.1 & 4.5 & 20.7 & 0.7 & \textbf{24.6} & 6.1 & 10.9 & -- & 10.3 & 0.7 \\
BioRED & \textbf{22.8} & 1.2 & 0.0 & 1.1 & 2.3 & 0.5 & -- & -- & -- & -- & -- & -- \\
GAP & \textbf{84.6} & 0.0 & 0.0 & 1.4 & 0.0 & 0.0 & -- & -- & -- & -- & -- & -- \\
LitBank & \textbf{24.3} & 0.0 & 0.0 & 0.1 & 0.0 & 0.0 & -- & -- & -- & -- & -- & -- \\
MAVEN-ERE & \textbf{7.0} & 0.0 & 0.0 & 0.0 & 0.0 & 0.0 & -- & -- & -- & -- & -- & -- \\
MeasEval & \textbf{58.5} & 2.5 & 8.9 & 0.0 & 1.0 & 0.8 & \textbf{15.3} & 1.7 & 6.2 & -- & 1.0 & 0.6 \\
Re-TACRED & \textbf{37.8} & 20.1 & 13.8 & 16.0 & 11.4 & 16.5$^\ddagger$ & -- & -- & -- & -- & -- & -- \\
\cmidrule(lr){1-13}
Average & \textbf{40.9} & 6.4 & 7.9 & 3.1 & 6.6 & 2.4$^\ddagger$ & \textbf{22.8} & 5.5 & 11.9 & -- & 7.1 & 0.6 \\
\bottomrule
\end{tabular}
\end{adjustbox}
\caption{RE with type names alone, reported as micro F1 (\%). Evaluation settings follow Table~\ref{tab:re}. Dashes mark unevaluated end-to-end conversions or GLiREL's unsupported end-to-end setting. $^\ddagger$ marks GoLLIE, which was trained on TACRED sentences reused by Re-TACRED. Best scores are \textbf{bold}.}
\label{tab:re_wo_desc}
\end{table*}

%% file: tables/ee_without_description.tex
\begin{table*}[t]
\centering
\small
\setlength{\tabcolsep}{3.5pt}
\begin{adjustbox}{max width=\textwidth}
\begin{tabular}{lrrrrrrrrrrrr}
\toprule
\multirow{2}{*}{Dataset} & \multicolumn{4}{c}{Arg.\ (gold triggers)} & \multicolumn{4}{c}{Trigger (end to end)} & \multicolumn{4}{c}{Arg.\ (end to end)} \\
\cmidrule(lr){2-5}\cmidrule(lr){6-9}\cmidrule(lr){10-13}
 & \textsc{CalOn} & GLiNER2 & InstructUIE & GoLLIE & \textsc{CalOn} & GLiNER2 & InstructUIE & GoLLIE & \textsc{CalOn} & GLiNER2 & InstructUIE & GoLLIE \\
\midrule
CASIE & 33.1 & 0.3 & \textbf{62.6}$^\ddagger$ & 2.1 & 31.8 & 0.6 & \textbf{67.6}$^\ddagger$ & 4.1 & 12.0 & 0.1 & \textbf{49.4}$^\ddagger$ & 1.4 \\
MLEE & \textbf{12.1} & 0.1 & 3.7 & 0.2 & \textbf{47.1} & 0.0 & 3.7 & 1.9 & \textbf{4.1} & 0.0 & 0.6 & 0.1 \\
M2E2 & 34.2 & 3.3 & \textbf{50.2} & 27.4 & 14.3 & 1.3 & 0.0 & \textbf{30.9} & 7.3 & 0.5 & 0.0 & \textbf{15.6} \\
MAILEX & \textbf{19.8} & 0.0 & 4.7 & 0.0 & \textbf{10.5} & 0.0 & 0.5 & 0.0 & \textbf{2.0} & 0.0 & 0.0 & 0.0 \\
SENTiVENT & \textbf{31.4} & 0.9 & 28.5 & 5.2 & \textbf{17.6} & 1.9 & 11.8 & 7.5 & \textbf{7.3} & 0.6 & 5.0 & 4.3 \\
SPEED++ mpox & \textbf{45.7} & 1.1 & 37.5 & 7.0 & \textbf{40.7} & 0.4 & 35.7 & 8.0 & \textbf{22.7} & 0.3 & 18.3 & 4.3 \\
SPEED++ zika & \textbf{53.7} & 0.0 & 45.5 & 17.9 & 41.1 & 0.0 & \textbf{49.4} & 20.8 & 25.5 & 0.0 & \textbf{26.5} & 12.3 \\
\cmidrule(lr){1-13}
Average & 32.9 & 0.8 & \textbf{33.2}$^\ddagger$ & 8.5 & \textbf{29.0} & 0.6 & 24.1$^\ddagger$ & 10.5 & 11.6 & 0.2 & \textbf{14.3}$^\ddagger$ & 5.4 \\
\bottomrule
\end{tabular}
\end{adjustbox}
\caption{EE with event and role names alone, reported as micro F1 (\%). Evaluation settings follow Table~\ref{tab:ee}. $^\ddagger$ marks scores for the evaluated InstructUIE checkpoint whose released training configuration includes CASIE. Best scores are \textbf{bold}.}
\label{tab:ee_wo_desc}
\end{table*}

%% file: tables/calibration.tex
\begin{table*}[t]
\centering
\small
\begin{tabular}{lrrrrrr}
\toprule
 & \multicolumn{3}{c}{F1 (\%)} & \multicolumn{3}{c}{ECE} \\
\cmidrule(lr){2-4} \cmidrule(lr){5-7}
Dataset & Fitted & +Corr. & +Corr.+$T$ & Fitted & +Corr. & +Corr.+$T$ \\
\midrule
\multicolumn{7}{l}{\textit{Entities}} \\
Validation (training sources) & 68.4 & 68.5 & 68.4 & 0.25 & 0.16 & \textbf{0.12} \\
\cmidrule(lr){1-7}
CrossNER AI & 66.1 & 66.0 & 65.8 & 0.26 & 0.21 & \textbf{0.17} \\
CrossNER Literature & 63.5 & 63.3 & 63.1 & 0.27 & 0.22 & \textbf{0.18} \\
CrossNER Music & 72.0 & 71.7 & 71.4 & 0.18 & 0.15 & \textbf{0.10} \\
CrossNER Politics & 73.0 & 72.8 & 72.6 & 0.23 & 0.18 & \textbf{0.15} \\
CrossNER Science & 64.4 & 64.4 & 64.3 & 0.21 & 0.17 & \textbf{0.12} \\
MIT Movie & 58.1 & 58.1 & 57.6 & 0.40 & 0.32 & \textbf{0.28} \\
MIT Restaurant & 66.5 & 66.5 & 66.2 & 0.31 & 0.23 & \textbf{0.20} \\
SciER & 70.9 & 70.7 & 70.7 & 0.24 & 0.15 & \textbf{0.12} \\
SciER (OOD) & 67.2 & 67.1 & 67.1 & 0.24 & 0.14 & \textbf{0.11} \\
GENTLE & 42.0 & 42.1 & 42.2 & 0.41 & 0.32 & \textbf{0.28} \\
BioRED & 72.1 & 72.0 & 71.9 & 0.21 & 0.13 & \textbf{0.10} \\
MeasEval & 30.6 & 30.3 & 29.8 & 0.62 & 0.49 & \textbf{0.45} \\
ai4privacy & 68.8 & 69.0 & 68.8 & 0.20 & 0.16 & \textbf{0.12} \\
spy legal & 53.2 & 53.1 & 53.0 & 0.59 & 0.56 & \textbf{0.54} \\
spy medical & 53.5 & 53.3 & 53.3 & 0.59 & 0.55 & \textbf{0.53} \\
\cmidrule(lr){1-7}
Average & 61.5 & 61.4 & 61.2 & 0.33 & 0.27 & \textbf{0.23} \\
\midrule
\multicolumn{7}{l}{\textit{Relations (gold entities)}} \\
Validation (training sources) & 82.8 & 82.8 & 82.9 & 0.13 & \textbf{0.01} & 0.03 \\
\cmidrule(lr){1-7}
CrossRE AI & 33.6 & 33.7 & 33.5 & 0.49 & 0.27 & \textbf{0.24} \\
CrossRE Literature & 48.2 & 48.4 & 48.2 & 0.35 & 0.12 & \textbf{0.09} \\
CrossRE Music & 51.6 & 51.8 & 51.4 & 0.26 & 0.08 & \textbf{0.06} \\
CrossRE News & 61.9 & 62.2 & 62.8 & \textbf{0.08} & 0.10 & 0.13 \\
CrossRE Politics & 40.4 & 40.4 & 40.2 & 0.28 & 0.13 & \textbf{0.11} \\
CrossRE Science & 36.8 & 36.9 & 36.6 & 0.33 & 0.15 & \textbf{0.11} \\
SciER & 43.4 & 43.8 & 43.1 & 0.19 & 0.13 & \textbf{0.12} \\
SciER (OOD) & 38.9 & 38.7 & 38.9 & 0.23 & \textbf{0.07} & 0.09 \\
ASTE 14lap & 75.6 & 75.6 & 75.4 & 0.13 & 0.09 & \textbf{0.09} \\
ASTE 14res & 86.3 & 86.3 & 86.3 & 0.07 & 0.04 & \textbf{0.04} \\
ASTE 15res & 83.9 & 83.8 & 83.8 & 0.11 & 0.09 & \textbf{0.07} \\
ASTE 16res & 86.3 & 86.2 & 86.4 & 0.10 & 0.08 & \textbf{0.07} \\
BioRED & 17.7 & 18.0 & 17.8 & 0.30 & 0.10 & \textbf{0.09} \\
GAP & 76.3 & 76.4 & 76.2 & 0.11 & \textbf{0.05} & 0.07 \\
LitBank & 16.8 & 17.3 & 16.1 & 0.40 & 0.04 & \textbf{0.03} \\
MAVEN-ERE & 2.0 & 2.2 & 2.1 & 0.45 & 0.15 & \textbf{0.13} \\
MeasEval & 57.9 & 58.2 & 57.9 & 0.29 & 0.06 & \textbf{0.05} \\
Re-TACRED & 38.4 & 38.7 & 38.2 & 0.26 & 0.16 & \textbf{0.12} \\
\cmidrule(lr){1-7}
Average & 49.8 & 49.9 & 49.7 & 0.25 & 0.11 & \textbf{0.09} \\
\midrule
\multicolumn{7}{l}{\textit{Event arguments (gold triggers)}} \\
CASIE & 37.9 & 38.1 & 38.2 & 0.59 & 0.48 & \textbf{0.45} \\
MLEE & 11.8 & 12.0 & 11.9 & 0.29 & 0.11 & \textbf{0.11} \\
M2E2 & 30.9 & 30.7 & 30.6 & 0.72 & 0.67 & \textbf{0.66} \\
MAILEX & 23.9 & 24.5 & 24.4 & 0.58 & 0.47 & \textbf{0.44} \\
SENTiVENT & 37.7 & 38.0 & 37.8 & 0.56 & 0.46 & \textbf{0.44} \\
SPEED++ mpox & 46.9 & 46.9 & 46.8 & 0.48 & 0.35 & \textbf{0.33} \\
SPEED++ zika & 56.8 & 57.0 & 56.6 & 0.37 & 0.31 & \textbf{0.30} \\
\cmidrule(lr){1-7}
Average & 35.1 & 35.3 & 35.2 & 0.51 & 0.41 & \textbf{0.39} \\
\bottomrule
\end{tabular}
\caption{F1 and ECE for \textsc{CalOn}. \textit{Fitted} uses the original calibration parameters without null-class correction. \textit{+Corr.} applies the correction and refits thresholds with $T = 1$. \textit{+Corr.+$T$} additionally refits temperature. Parameters are selected on validation data from the training sources, using REBEL for relations and event arguments, and remain fixed on evaluation datasets. Relations use gold entities, and event arguments use gold triggers. ECE uses 15 equal-width bins over returned predictions and the probability of the predicted type or role. Lowest ECE values are \textbf{bold}.}
\label{tab:calibration}
\end{table*}

%% file: tables/models.tex
\begin{table*}[t]
\centering
\small
\begin{adjustbox}{max width=\textwidth}
\begin{tabular}{lllll}
\toprule
\textbf{Model} & \textbf{Architecture} & \textbf{Model link} & \textbf{License} & \textbf{Release} \\
\midrule

\rowcolor{IE!30}
Qwen3-0.6B & Dense &
\href{https://huggingface.co/Qwen/Qwen3-0.6B}{Qwen/Qwen3-0.6B} &
Apache 2.0 & Apr 2025 \\

\rowcolor{YW!80}
GLiNER2-large & Dense &
\href{https://huggingface.co/fastino/gliner2-large-v1}{fastino/gliner2-large-v1} &
Apache 2.0 & Jul 2025 \\

\rowcolor{YW!80}
GLiFormer-large & Dense &
\href{https://huggingface.co/knowledgator/gliformer-large-v1}{knowledgator/gliformer-large-v1} &
Apache 2.0 & Sep 2026 \\

\rowcolor{YW!80}
GLiREL-large & Dense &
\href{https://huggingface.co/jackboyla/glirel-large-v0}{jackboyla/glirel-large-v0} &
CC BY-NC-SA 4.0 & Jan 2025 \\

\rowcolor{YW!80}
GNER-LLaMA-7B & Dense &
\href{https://huggingface.co/dyyyyyyyy/GNER-LLaMA-7B}{dyyyyyyyy/GNER-LLaMA-7B} &
Apache 2.0 & Feb 2024 \\

\rowcolor{YW!80}
InstructUIE & Dense &
\href{https://huggingface.co/ZWK/InstructUIE}{ZWK/InstructUIE} &
OpenRAIL & May 2023 \\

\rowcolor{YW!80}
GoLLIE-7B & Dense &
\href{https://huggingface.co/HiTZ/GoLLIE-7B}{HiTZ/GoLLIE-7B} &
Llama 2 Community & Sep 2023 \\

\rowcolor{ST!45}
Qwen3.8-27B & Dense &
\href{https://huggingface.co/Qwen/Qwen3.8-27B}{Qwen/Qwen3.8-27B} &
Apache 2.0 & Aug 2026 \\

\rowcolor{ST!45}
gpt-oss-120b & MoE &
\href{https://huggingface.co/openai/gpt-oss-120b}{openai/gpt-oss-120b} &
Apache 2.0 & Aug 2025 \\

\rowcolor{ST!45}
Nemotron-3-Nano-Omni-30B-A3B & MoE &
\href{https://huggingface.co/nvidia/Nemotron-3-Nano-Omni-30B-A3B-Reasoning-BF16}{nvidia/Nemotron-3-Nano-Omni-30B-A3B-Reasoning-BF16} &
NVIDIA Open Model & Apr 2026 \\

\rowcolor{ST!45}
Phi-4 & Dense &
\href{https://huggingface.co/microsoft/phi-4}{microsoft/phi-4} &
MIT & Dec 2024 \\

\rowcolor{ST!45}
Mistral-Small-24B & Dense &
\href{https://huggingface.co/mistralai/Mistral-Small-24B-Instruct-2501}{mistralai/Mistral-Small-24B-Instruct-2501} &
Apache 2.0 & Jan 2025 \\

\rowcolor{WP!70}
all-mpnet-base-v2 & Dense &
\href{https://huggingface.co/sentence-transformers/all-mpnet-base-v2}{sentence-transformers/all-mpnet-base-v2} &
Apache 2.0 & Mar 2022 \\

\bottomrule
\end{tabular}
\end{adjustbox}
\caption{Model checkpoints used in this work, with their architecture, Hugging Face links, reported licenses, and release months. 
% The backbone is shown in blue, baselines in pink, models used for data generation and annotation in sand, and the sentence encoder in lilac.
}
\label{tab:models}
\end{table*}

%% file: tables/datasets.tex
\begin{table*}[t]
\centering
\resizebox{\linewidth}{!}{
\begin{tabular}{lllrr}
\toprule
\textbf{Dataset} & \textbf{Task} & \textbf{Domain and genre} & \textbf{\# Types} & \textbf{\# Examples} \\
\midrule
\rowcolor{IE!30} CrossNER AI~\cite{liu2021crossner} & Entities & AI, encyclopedia & 14 & 431 \\
\rowcolor{IE!30} CrossNER Literature~\cite{liu2021crossner} & Entities & Literature, encyclopedia & 12 & 416 \\
\rowcolor{IE!30} CrossNER Music~\cite{liu2021crossner} & Entities & Music, encyclopedia & 13 & 465 \\
\rowcolor{IE!30} CrossNER Politics~\cite{liu2021crossner} & Entities & Politics, encyclopedia & 9 & 651 \\
\rowcolor{IE!30} CrossNER Science~\cite{liu2021crossner} & Entities & Science, encyclopedia & 17 & 543 \\
\rowcolor{IE!30} MIT Movie~\cite{liu2013asgard} & Entities & Film, search queries & 12 & 1,953 \\
\rowcolor{IE!30} MIT Restaurant~\cite{liu2013asgard} & Entities & Food, search queries & 8 & 1,521 \\
\rowcolor{IE!30} SciER \cite{zhang-etal-2024-scier} & Entities & AI, scientific papers & 3 & 854 \\
\rowcolor{IE!30} SciER (OOD) \cite{zhang-etal-2024-scier} & Entities & Science, scientific papers & 3 & 580 \\
\rowcolor{IE!30} GENTLE \cite{aoyama-etal-2023-gentle} & Entities & Mixed, unusual genres & 10 & 1,334 \\
\rowcolor{IE!30} BioRED \cite{luo-etal-2022-biored} & Entities & Biomedicine, abstracts & 6 & 118 \\
\rowcolor{IE!30} MeasEval \cite{harper-etal-2021-semeval} & Entities & Science, scientific papers & 4 & 149 \\
\rowcolor{IE!30} ai4privacy \cite{ai4privacy_open_pii_masking_500k} & Entities & Personal information, forms and messages & 25 & 3,000 \\
\rowcolor{IE!30} spy legal \cite{savkin-etal-2025-spy} & Entities & Personal information, legal documents & 7 & 12,208 \\
\rowcolor{IE!30} spy medical \cite{savkin-etal-2025-spy} & Entities & Personal information, medical documents & 7 & 11,785 \\
\rowcolor{YW!80} CrossRE AI~\cite{bassignana-plank-2022-crossre} & Relations & AI, encyclopedia & 17 & 431 \\
\rowcolor{YW!80} CrossRE Literature~\cite{bassignana-plank-2022-crossre} & Relations & Literature, encyclopedia & 17 & 416 \\
\rowcolor{YW!80} CrossRE Music~\cite{bassignana-plank-2022-crossre} & Relations & Music, encyclopedia & 17 & 399 \\
\rowcolor{YW!80} CrossRE News~\cite{bassignana-plank-2022-crossre} & Relations & General, news & 17 & 400 \\
\rowcolor{YW!80} CrossRE Politics~\cite{bassignana-plank-2022-crossre} & Relations & Politics, encyclopedia & 17 & 400 \\
\rowcolor{YW!80} CrossRE Science~\cite{bassignana-plank-2022-crossre} & Relations & Science, encyclopedia & 17 & 400 \\
\rowcolor{YW!80} SciER \cite{zhang-etal-2024-scier} & Relations & AI, scientific papers & 9 & 854 \\
\rowcolor{YW!80} SciER (OOD) \cite{zhang-etal-2024-scier} & Relations & Science, scientific papers & 9 & 580 \\
\rowcolor{YW!80} ASTE 14lap \cite{xu-etal-2020-position} & Relations & Laptops, reviews & 3 & 328 \\
\rowcolor{YW!80} ASTE 14res \cite{xu-etal-2020-position} & Relations & Restaurants, reviews & 3 & 492 \\
\rowcolor{YW!80} ASTE 15res \cite{xu-etal-2020-position} & Relations & Restaurants, reviews & 3 & 322 \\
\rowcolor{YW!80} ASTE 16res \cite{xu-etal-2020-position} & Relations & Restaurants, reviews & 3 & 326 \\
\rowcolor{YW!80} BioRED \cite{luo-etal-2022-biored} & Relations & Biomedicine, abstracts & 8 & 140 \\
\rowcolor{YW!80} GAP \cite{webster-etal-2018-mind} & Relations (pronouns) & People, encyclopedia & 1 & 2,000 \\
\rowcolor{YW!80} LitBank \cite{bamman-etal-2020-annotated} & Relations (coreference) & Fiction, novels & 1 & 1,388 \\
\rowcolor{YW!80} MAVEN-ERE \cite{wang-etal-2022-maven} & Relations (events) & History, encyclopedia & 10 & 1,250 \\
\rowcolor{YW!80} MeasEval \cite{harper-etal-2021-semeval} & Relations & Science, scientific papers & 3 & 149 \\
\rowcolor{YW!80} Re-TACRED \cite{stoica2021re} & Relations & People and organisations, news and web & 38 & 13,418 \\
\rowcolor{ST!45} CASIE \cite{satyapanich2020casie} & Events & Cybersecurity, news & 5 & 1,492 \\
\rowcolor{ST!45} MLEE \cite{pyysalo-etal-2012-mlee} & Events & Biomedicine, abstracts & 22 & 98 \\
\rowcolor{ST!45} M2E2 \cite{li-etal-2020-cross} & Events & Politics and conflict, news & 8 & 5,996 \\
\rowcolor{ST!45} MAILEX \cite{srivastava-etal-2023-mailex} & Events & Business, emails & 10 & 406 \\
\rowcolor{ST!45} SENTiVENT \cite{jacobs2022sentivent} & Events & Finance, news & 18 & 727 \\
\rowcolor{ST!45} SPEED++ mpox \cite{parekh-etal-2024-speed} & Events & Epidemics, social media & 7 & 286 \\
\rowcolor{ST!45} SPEED++ zika \cite{parekh-etal-2024-speed} & Events & Epidemics, social media & 7 & 299 \\
\bottomrule
\end{tabular}
}
\caption{The 40 evaluation benchmarks with their tasks, domains, genres, type counts, and numbers of converted evaluation inputs. Types are entity, relation, or event types according to the task. Inputs may be sentences, passages, or document chunks. All scores use micro F1. 
% Entity benchmarks are shown in blue, relation benchmarks in pink, and event benchmarks in sand.
}
\label{tab:datasets}
\end{table*}

%% file: tables/train_datasets.tex
\begin{table*}[t]
\centering
\resizebox{\linewidth}{!}{
\begin{tabular}{llllrr}
\toprule
\textbf{Dataset} & \textbf{Task} & \textbf{Domain and genre} & \textbf{Annotation} & \textbf{\# Types} & \textbf{\# Examples} \\
\midrule
\rowcolor{IE!30} Few-NERD~\cite{ding-etal-2021-nerd} & Entities & General, encyclopedia & Human & 74 & 125,839 \\
\rowcolor{IE!30} MASSIVE \cite{fitzgerald-etal-2023-massive} & Entities (slots) & Personal assistant, voice commands & Human & 55 & 11,062 \\
\rowcolor{IE!30} FiNERWeb~\cite{golde-etal-2026-finerweb} & Entities & General, web pages & LLM (released) & 3,895 & 16,798 \\
\rowcolor{IE!30} Pile-NER~\cite{zhou2024universalner} & Entities & General, mixed text & LLM (released) & 268,736 & 73,395 \\
\rowcolor{YW!80} REBEL \cite{huguet-cabot-navigli-2021-rebel-relation} & Entities and relations & General, encyclopedia & LLM (ours) & 45,074 & 399,983 \\
\rowcolor{ST!45} MAVEN-Arg \cite{wang-etal-2024-maven} & Events & General, encyclopedia & Human & 162 & 12,889 \\
\rowcolor{ST!45} GENEVA \cite{parekh-etal-2023-geneva} & Events & General, mixed & Human & 115 & 6,112 \\
\rowcolor{ST!45} RAMS \cite{ebner-etal-2020-multi} & Events & Politics and conflict, news & Human & 139 & 15,786 \\
\rowcolor{ST!45} PHEE \cite{sun-etal-2022-phee} & Events & Pharmacovigilance, case reports & Human & 2 & 8,537 \\
\rowcolor{ST!45} CC-News & Events & General, news & LLM (ours) & 45,543 & 295,343 \\
\bottomrule
\end{tabular}
}
\caption{Training sources using real texts. \textit{Human} denotes the original human annotations, \textit{LLM (released)} denotes released LLM annotations, and \textit{LLM (ours)} denotes annotations produced by our pipeline. Type counts include names with at least one training annotation, counting relation types for REBEL and event types for event sources. Example counts refer to converted training instances, including trigger and argument instances derived from event passages. Synthetic sources are described in Table~\ref{tab:synth_grid} and Figure~\ref{fig:prompts}.}
\label{tab:train_datasets}
\end{table*}

%% file: tables/synth_grid.tex
\begin{table*}[t]
\centering
\small
\begin{tabular}{p{0.12\textwidth}p{0.82\textwidth}}
\toprule
\rowcolor{IE!30}
\textbf{Genres} &
email; text message conversation; internal business report; meeting notes; social media post; product review; short news brief; customer support ticket; invoice or order confirmation; job posting; contract clause; clinical note; research paper abstract; recipe; travel itinerary; forum discussion post; press release; resume summary; encyclopedia-style paragraph; podcast transcript excerpt; spoken request to a voice assistant; search query log; legal contract summary; government notice; patent abstract; wiki talk page comment; product specification sheet; hospital discharge summary; sports match report; obituary; museum exhibit label; travel guide paragraph; restaurant menu description; movie synopsis; tweet thread; answer on a question-and-answer site; tutorial step; software changelog entry; science news article; court judgment excerpt \\
\midrule
\rowcolor{YW!80}
\textbf{Domains} &
finance and banking; healthcare; pharmaceuticals; law and courts; politics and elections; government and public policy; military and defense; sports; music; film and television; literature and publishing; visual arts; video games; fashion; food and restaurants; travel and tourism; hotels and hospitality; real estate; construction and architecture; automotive; aviation; shipping and logistics; retail and e-commerce; consumer electronics; telecommunications; software engineering; cybersecurity; artificial intelligence research; data science; physics; chemistry; biology and genetics; astronomy and space; earth science and climate; agriculture; energy and utilities; oil and gas; mining; manufacturing; human resources; education; academia and universities; religion; history; archaeology; philosophy; linguistics; journalism and media; advertising and marketing; insurance; cryptocurrency and blockchain; nonprofit and charity; environmental conservation; veterinary and animals; sports betting and gaming; social work; public transport; weddings and events; fitness and wellness; parenting and family; ophthalmology; dentistry; mental health; nutrition; tennis; association football; basketball; baseball; cricket; motorsport; cycling; the Olympic Games; chess; board games; anime and manga; theatre; dance; photography; sculpture; museums; classical antiquity; medieval history; military history; genealogy; urban planning; maritime industry; railways; space industry; robotics; semiconductors; cloud computing; open-source software; mobile apps; social networks; e-sports; tax law; intellectual property; immigration; criminal justice; human rights; international relations; diplomacy; local government; labor unions; supply chains; wine and beverages; coffee and tea; gardening; pets; zoology; botany; marine biology; ecology; geology; meteorology; oceanography; epidemiology; neuroscience; materials science; mathematics \\
\bottomrule
\end{tabular}
\caption{Genres and domains used for synthetic generation. Each request combines one genre with one domain and asks the generator to produce a text and an extraction schema.}
\label{tab:synth_grid}
\end{table*}

%% file: tables/prompts.tex
\begin{figure*}[p]
    \begin{tcolorbox}[
        enhanced,
        colback=white,
        colframe=black,
        arc=3mm,
        boxrule=0.5pt,
        fontupper=\footnotesize,
        title=Prompt Template (Entities).
    ]
    Write one realistic \{genre\} from the domain of \{domain\}, about \{length\} words long, in English. Make it specific and natural, with concrete names, places, numbers and dates where the genre calls for them.

    \vspace{1ex}
    Then design an entity-extraction schema for this text: 6 to 12 entity types a user working in \{domain\} would want to extract from texts like it. Include at least two pairs of closely related types that must be told apart (for example a general type and a more specific one, or two similar roles). Give every type a one-sentence definition that would apply to other texts too.
    Annotate every mention of every schema type exactly as it appears in the text; schema types that do not occur get no entries.
    Finally list 3 to 5 further entity types that are relevant to \{domain\} but do not occur anywhere in this text, each with a one-sentence definition.

    \vspace{1ex}
    Return JSON only: \texttt{\{"text": "...", "types": \{"type name": "definition"\}, "entities": [\{"text": "...", "type": "..."\}], "absent\_types": \{"type name": "definition"\}\}}
    \end{tcolorbox}
    \vspace{1.5ex}

    \begin{tcolorbox}[
        enhanced,
        colback=white,
        colframe=black,
        arc=3mm,
        boxrule=0.5pt,
        fontupper=\footnotesize,
        title=Prompt Template (Relations).
    ]
    Write one realistic \{genre\} from the domain of \{domain\}, about \{length\} words long, in English. Make it specific and natural, with concrete names, places, organisations, products, numbers and dates where the genre calls for them, and with several entities that are related to each other.

    \vspace{1ex}
    Then design an extraction schema for this text:\\
    - 6 to 12 entity types a user working in \{domain\} would want to extract, each with a one-sentence definition that would apply to other texts too;\\
    - 5 to 10 directed relation types between those entities (from a head entity to a tail entity), each with a one-sentence definition. Include at least two pairs of closely related relation types that must be told apart (for example ``founded'' and ``employed by'', or a relation and its reverse direction).\\
    Annotate every mention of every entity type exactly as it appears in the text, and every relation of a schema type that the text states between two of those mentions (head and tail copied exactly as they appear). Schema types that do not occur get no entries.
    Finally list 3 to 5 further relation types that would be plausible between these entities but are not stated in the text, each with a one-sentence definition.

    \vspace{1ex}
    Return JSON only: \texttt{\{"text": "...", "entity\_types": \{"type": "definition"\}, "entities": [\{"text": "...", "type": "..."\}], "relation\_types": \{"relation": "definition"\}, "relations": [\{"head": "...", "relation": "...", "tail": "..."\}], "absent\_relations": \{"relation": "definition"\}\}}
    \end{tcolorbox}
    \vspace{1.5ex}

    \begin{tcolorbox}[
        enhanced,
        colback=white,
        colframe=black,
        arc=3mm,
        boxrule=0.5pt,
        fontupper=\footnotesize,
        title=Prompt Template (Events).
    ]
    Write one realistic \{genre\} from the domain of \{domain\}, about \{length\} words long, in English. Make it specific and natural, with concrete names, places, numbers and dates where the genre calls for them, and let it report several events (things that happen or are done), with their participants, times, places and other details.

    \vspace{1ex}
    Then design an event schema for this text: 3 to 6 event types a user working in \{domain\} would want to extract, each with a one-sentence definition that would apply to other texts too, and for each event type 2 to 6 argument roles (the participants and attributes of such an event), each with a one-sentence definition. Include at least one pair of closely related event types that must be told apart.
    Annotate every event of every schema type in the text: its type, its trigger (the single word, occasionally a short phrase, that most directly expresses the event, copied exactly) and its arguments (role and the span filling it, copied exactly as it appears; only roles of that event type). Event types of the schema that do not occur get no events.
    Finally list 2 to 4 further event types that are relevant to \{domain\} but do not occur anywhere in this text, each with a one-sentence definition.

    \vspace{1ex}
    Return JSON only: \texttt{\{"text": "...", "event\_types": \{"type": \{"definition": "...", "roles": \{"role": "definition"\}\}\}, "events": [\{"type": "...", "trigger": "...", "arguments": [\{"role": "...", "text": "..."\}]\}], "absent\_event\_types": \{"type": "definition"\}\}}
    \end{tcolorbox}
    \caption{Generation prompts for entities, relations, and events. The genre and domain values come from Table~\ref{tab:synth_grid}. Requested text lengths are 30, 60, 120, 200, or 300 words. Types listed as absent provide explicit negative supervision after annotation checking.}
    \label{fig:prompts}
\end{figure*}